\documentclass[10pt,twocolumn,letterpaper]{article}

\usepackage[pagenumbers]{cvpr} % To force page numbers, e.g. for an arXiv version

\definecolor{cvprblue}{rgb}{0.21,0.49,0.74}
\usepackage[pagebackref,breaklinks,colorlinks,allcolors=cvprblue]{hyperref}

\usepackage{algorithm}
\usepackage{algorithmic}

\usepackage{multirow}
\usepackage{makecell}
\usepackage{graphicx}
\usepackage{colortbl} % provide colorcell
\usepackage{arydshln} % provide \hdashline  \cdashline
\usepackage{booktabs} % provides toprule
\usepackage{array}
\usepackage{diagbox}
\usepackage{bbding}

\usepackage{cuted}
\usepackage{hyperref}

\usepackage{overpic}
\usepackage{caption}

\usepackage{amsmath}
\usepackage{amssymb}
\usepackage{bm}
\usepackage{oz}

\usepackage{xcolor}
\newcommand{\foretab}{\rowcolor{gray!10}}

\usepackage[accsupp]{axessibility}

\def\paperID{7181} % *** Enter the Paper ID here
\def\confName{CVPR}
\def\confYear{2025}

\title{Diff-Palm: Realistic Palmprint Generation with Polynomial Creases \\
and Intra-Class Variation Controllable Diffusion Models}

\author{
    Jianlong Jin\textsuperscript{\rm 1}\footnotemark[1],
    Chenglong Zhao\textsuperscript{\rm 2}\footnotemark[1],
    Ruixin Zhang\textsuperscript{\rm 2},
    Sheng Shang\textsuperscript{\rm 1},
    Jianqing Xu\textsuperscript{\rm 2}, 
    Jingyun Zhang\textsuperscript{\rm 3}, \\
    ShaoMing Wang\textsuperscript{\rm 3}, 
    Yang Zhao\textsuperscript{\rm 1},
    Shouhong Ding\textsuperscript{\rm 2}\footnotemark[2],
    Wei Jia\textsuperscript{\rm 1}\footnotemark[2],
    Yunsheng Wu\textsuperscript{\rm 2}\\
    {\small \textsuperscript{\rm 1} Hefei University of Technology, \textsuperscript{\rm 2} Tencent Youtu Lab, \textsuperscript{\rm 3} Tencent WeChat Pay Lab33 }\\
    {\tt \small jianlong@mail.hfut.edu.cn, lornezhao@tencent.com}
 }

\begin{document}

\twocolumn[{%
\renewcommand\twocolumn[1][]{#1}%
\maketitle
\begin{center}
    \vspace{-3mm}
    \captionsetup{type=figure}
    \includegraphics[width=.95\linewidth]{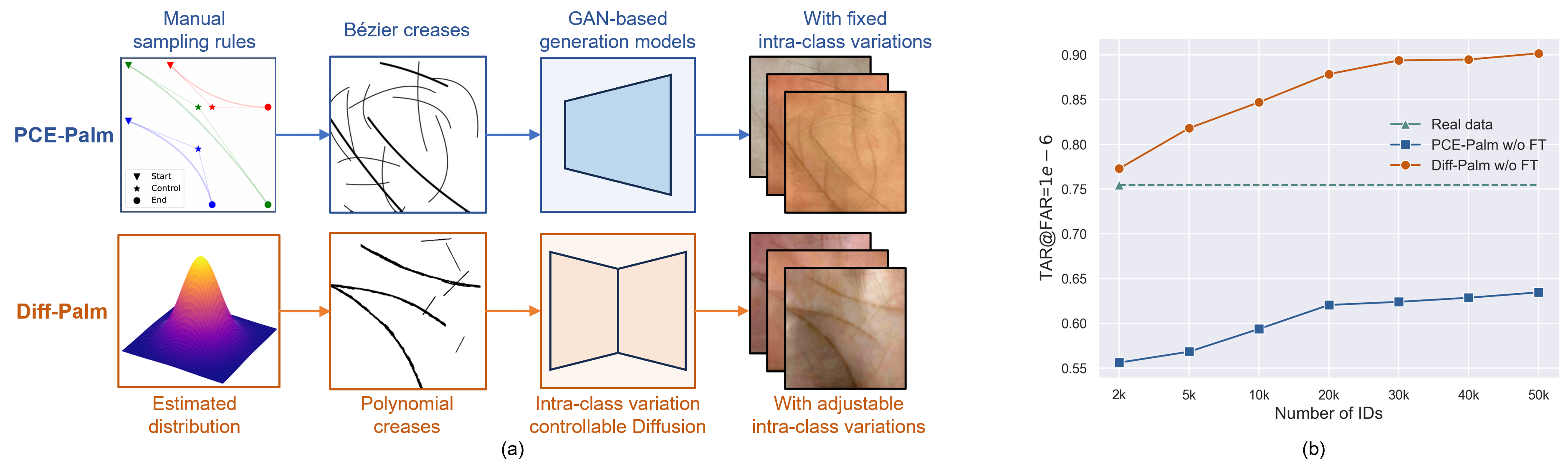}
    \vspace{-3mm}
    \caption{Comparison between PCE-Palm \cite{jin2024pce} and the proposed Diff-Palm. (a) PCE-Palm adopts conditional GAN with  B\'ezier creases \cite{zhao2022bezierpalm} as input to generate palmprint datasets. Diff-Palm introduces a polynomial crease and a novel diffusion model for synthesizing datasets with adjustable intra-class variations. (b) The average performance of recognition models, trained on three types of datasets (real data,  PCE-Palm generated, and Diff-Palm generated) and evaluated on five public datasets. (FT: Fine-tune) Code: \href{https://github.com/Ukuer/Diff-Palm}{https://github.com/Ukuer/Diff-Palm}}
    \label{fig:motivation}
\end{center}

}]

\renewcommand{\thefootnote}{\fnsymbol{footnote}} 

\footnotetext[1]{ Equal contribution. \footnotemark[2] Corresponding authors.}

\begin{abstract}
Palmprint recognition is significantly limited by the lack of large-scale publicly available datasets. 
Previous methods have adopted B\'ezier curves to simulate the palm creases, which then serve as input for conditional GANs to generate realistic palmprints.
However, without employing real data fine-tuning, the performance of the recognition model trained on these synthetic datasets would drastically decline, indicating a large gap between generated and real palmprints.
This is primarily due to the utilization of an inaccurate palm crease representation and challenges in balancing intra-class variation with identity consistency.
To address this, we introduce a polynomial-based palm crease representation that provides a new palm crease generation mechanism more closely aligned with the real distribution. 
We also propose the palm creases conditioned diffusion model with a novel intra-class variation control method.
By applying our proposed $K$-step noise-sharing sampling, we are able to synthesize palmprint datasets with large intra-class variation and high identity consistency.
Experimental results show that, for the first time, recognition models trained solely on our synthetic datasets, without any fine-tuning, outperform those trained on real datasets.
Furthermore, our approach achieves superior recognition performance as the number of generated identities increases.
\end{abstract}

\section{Introduction}
Palmprint recognition has gained widespread interest for its highly discriminative, user-friendly, and privacy-friendly nature \cite{fei2018feature}.
However, its further development is severely limited by the lack of large-scale public datasets.
With the remarkable results achieved by deep generative models, such as Generative Adversarial Networks (GANs) \cite{goodfellow2020generative} and Diffusion models \cite{ho2020denoising}, generating datasets for recognition tasks is a promising and valuable approach to replace the collection of large-scale real datasets \cite{joshi2024synthetic}.

Although existing palmprint generation methods, including RPG-Palm \cite{shen2023rpg} and PCE-Palm \cite{jin2024pce} have obtained impressive results, they still rely on fine-tuning with real data to obtain good recognition performance.
Specifically, they generate large-scale data to pre-train the recognition models and then fine-tune them with real data. 
However, without fine-tuning, the recognition model's performance suffers from significant degradation, indicating a large gap between the generated data and real data.

We have identified two principal factors contributing to the observed issue.
The first factor is the \textbf{inaccurate representation of palm creases}.
Existing methods \cite{shen2023rpg, jin2024pce} employ manually designed B\'ezier curves \cite{zhao2022bezierpalm} to simulate palm creases.
However, these curves significantly differ from the actual patterns found in genuine palmprints, resulting in a substantial divergence between synthetic and real palmprint data.
Furthermore, the use of merely three control points per B\'ezier curve restricts its expressiveness.
The second factor is the \textbf{challenges in balancing intra-class variation with identity consistency}.
In the context of palmprint image generation, these methods \cite{shen2023rpg, jin2024pce} employ conditional GANs and simulate intra-class variations of palmprints by adding a simple random noise. 
This strategy, however, yields datasets with large intra-class variation and low identity consistency compared to real datasets, thereby leading to diminished performance.

To address the aforementioned issues, we propose a novel framework to generate realistic palmprint datasets with large intra-class variation and high identity consistency, which achieves superior performance compared to the existing SOTA method, i.e., PCE-Palm \cite{jin2024pce}, as shown in Fig.\ref{fig:motivation}. 
Specifically, we utilize polynomial curves to describe palm creases, termed polynomial creases. 
To achieve a more accurate and expressive representation,
we employ fourth-order polynomial curves parameterized by five coefficients, which substantially expands the sampling space.
Based on the statistical analysis of real palm creases, we estimate a multivariate Gaussian distribution for the coefficients of polynomial curves. 
Subsequently, we develop a new sampling mechanism that obtains coefficients from the estimated Gaussian distribution to synthesize polynomial palm creases. 
This approach effectively reduces the discrepancy between synthetic and real palm creases.

Furthermore, we introduce a novel intra-class variation controllable diffusion model, designed to generate palmprints datasets that exhibit substantial variation while preserving high identity consistency.
Diffusion models \cite{ho2020denoising} synthesize images via a denoising sampling process. However, the inherent randomness of this process typically results in uncontrollable variation within the same identity class, potentially compromising identity consistency. 
This issue can lead to generated datasets with excessive intra-class variation and diminished identity consistency.
To tackle this problem, we propose a simple yet effective sampling method, termed
$K$-step noise-sharing sampling.
This method enhances the diffusion model by employing shared noise across samples within the same identity during the sampling process, instead of distinct noise sequences. 
By adjusting the parameter $K$, we are able to generate datasets with a spectrum of intra-class variations. 
Notably, the proposed $K$-step noise-sharing sampling can be used as a plug-and-play module applied in other diffusion models.

In our experimental setup, 
we train recognition models with synthetic datasets, and evaluate them directly on publicly available datasets without employing any real data fine-tuning.
Our contributions are as follows:

\begin{itemize}

\item[$\bullet$] 
We introduce a more realistic palm crease representation based on polynomial curves, 
which not only exhibits high expressive capability but also ensures that generated palm creases align with the distribution of real ones.

\item[$\bullet$] 
We propose a novel intra-class variation controllable diffusion model with a $K$-step noise-sharing sampling.
It utilizes palm creases as identity conditions, enabling the synthesis of palmprint datasets that exhibit large intra-class variation while preserving high identity consistency.

\item[$\bullet$] 
Through extensive open-set experiments,
we demonstrate for the first time that recognition models trained solely on synthetic datasets generated by our method outperform those trained on real datasets, without the benefit of real data fine-tuning. 
Moreover, our approach shows a significant improvement in performance as the number of generated identities increases.

\end{itemize}

\section{Related Work}

\subsection{Data Generation for Biometric Recognition}
With advancements in generation methods \cite{cao2024survey}, various models have been employed to create synthetic samples in the field of biometrics, such as 
face generation \cite{deng2020disentangled, nguyen2019hologan, qiu2021synface, fu2019dual, fu2021dvg, kim2023dcface, yan2024dialoguenerf}, 
and fingerprint synthesis~\cite{bahmani2021high, shoshan2024fpgan, engelsma2022printsgan}.
For palmprint generation, 
B\'ezierpalm \cite{zhao2022bezierpalm} firstly utilizes B\'ezier curves to represent palm creases and generates new identities by sampling random curves.
Subsequently, both RPG-Palm \cite{shen2023rpg} and PCE-Palm \cite{jin2024pce} employ B\'ezier curves as an identity condition and GAN-based generative models for palmprint generation.
Although these methods can produce visually realistic palmprint images, they still need to be fine-tuned with real data to achieve satisfactory results.

\subsection{Conditional Diffusion Models}
Diffusion models have achieved remarkable performance in numerous visual tasks \cite{croitoru2023diffusion, Zheng2024ResearchOD}. 
To enable controllable generation, two main conditional control mechanisms are employed.
The first is the embedding-based conditional approach \cite{rombach2022high, koley2024handle}, 
where the embedding is typically derived from the output of a pre-trained model.
This extracted embedding is then incorporated into the UNet network architecture through a cross-attention mechanism.
The second is the channel-based conditional approach \cite{saharia2022palette, mou2024t2i},
where conditional images are concatenated with the diffused image through channels. 
For our research, we utilize the channel-based conditional diffusion model as our baseline, 
as it ensures a strong correspondence between conditional images and generated images.
Furthermore, we draw comparisons with the embedding-based conditional model, IDiff-Face \cite{boutros2023idiff}, an identity-conditioned face generation method with impressive recognition performance.

\subsection{Recognition Methods}
Deep learning has significantly advanced biometric recognition, especially in facial recognition, where margin-based methods like CosFace \cite{wang2018cosface} and ArcFace \cite{deng2019arcface} have shown exceptional performance. 
Similarly, in palmprint recognition, several deep learning methods have been proposed~\cite{genovese2019palmnet, zhong2019centralized,jia2022eepnet,yang2023co3net,yang2023ccnet, fan2024novel}, many of which introduce refined networks companied by margin-based loss. 
Thus, this paper adopts ArcFace as the palmprint recognition baseline to compare different generation methods, as in PCE-Palm \cite{jin2024pce}.

\section{Methods}
In this section, we first introduce a polynomial representation for palm creases, which synthesize pseudo palm creases as identity conditions for generating realistic palmprints. 
We then propose an intra-class variation controllable diffusion model, 
which includes palm creases conditioned diffusion and a $K$-step noise-sharing sampling mechanism. 
Utilizing these components, we are able to generate palmprint datasets with adjustable levels of intra-class variations.
\subsection{Polynomial Representation for Palm Creases}

\begin{figure}[tb]
   \centering
   \center{\includegraphics[width=.95\linewidth]  {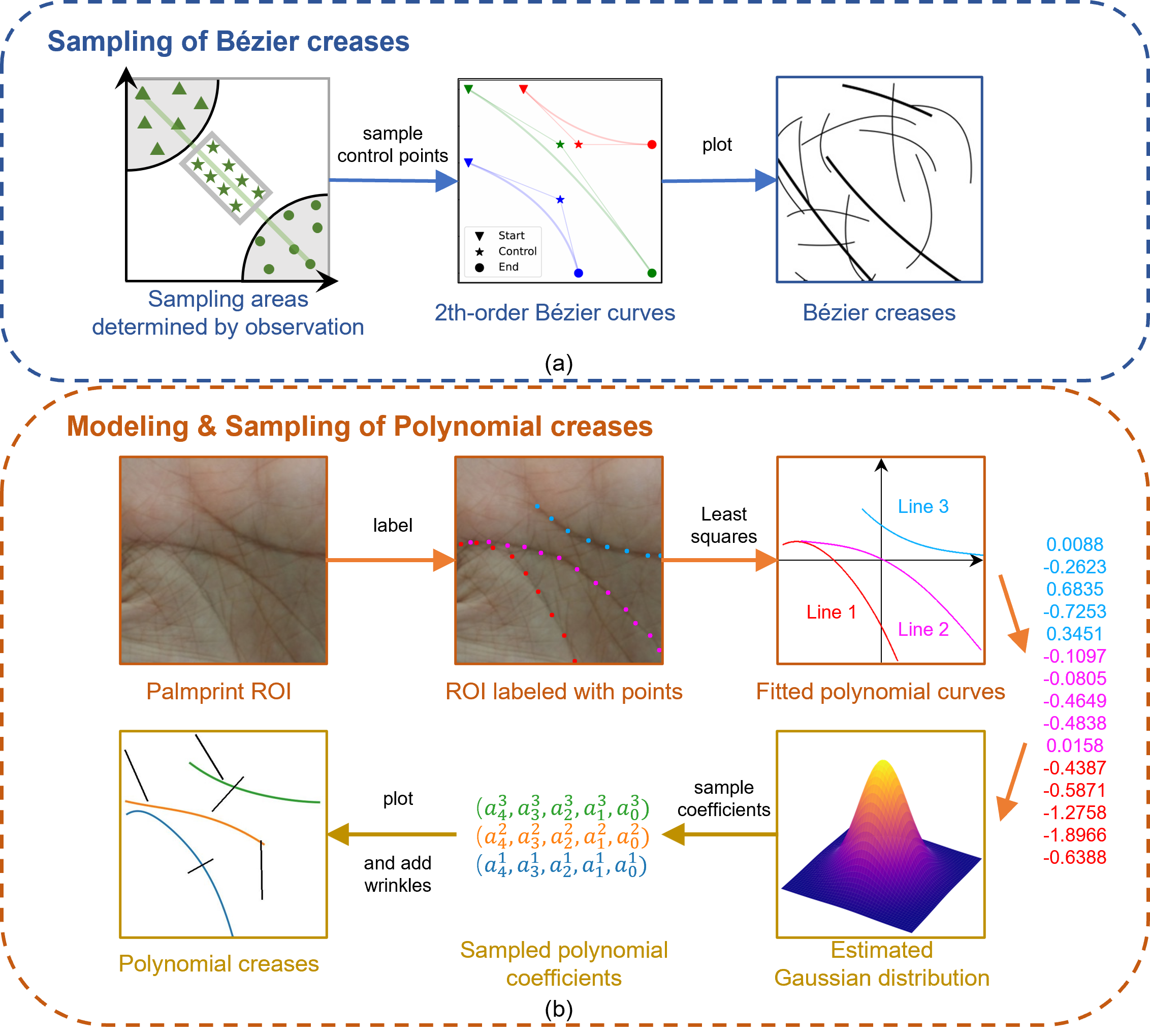 }}
   \caption{Comparison of (a) B\'ezier creases \cite{zhao2022bezierpalm} and (b) proposed polynomial creases.
   \vspace{-3mm}
   }\label{fig:poly}
   
\end{figure}

Current palmprint generation models \cite{shen2023rpg, jin2024pce} rely on palm creases as identity control conditions, making palm crease representation a crucial task for palmprint generation. 
These models all employ second-order B\'ezier curves to represent palm creases \cite{zhao2022bezierpalm}, 
as follows,
\begin{equation}
    \mathbf B(t) = (1 - t)^2 \mathbf P_s + 2t(1 - t) \mathbf P_c + t^2 \mathbf P_e,    
\end{equation}
where $\mathbf P_s, \mathbf P_c, \mathbf P_e$ represent starting point, control point, and ending point respectively, which are randomly sampled within a predefined artificial area range, and $t$ is the parameter ranging from $0$ to $1$, determining the position along the curve, as shown in Fig.\ref{fig:poly} (a).
However, although existing generation-based methods \cite{shen2023rpg, jin2024pce} can produce visually realistic palmprint images, the distribution of simulated creases is quite different from that of real palm creases, leading to the requirement of fine-tuning on real data. 

To overcome this limitation, this paper focuses on enhancing palm crease representation to align more closely with the distribution of real palm creases. Consequently, we propose a polynomial representation for palm creases, and derive the estimated distribution of the representation parameters through statistical analysis.
The proposed polynomial representation is defined as follows,
\begin{equation}
     y^i=a_4^i x^4+a_3^i x^3+a_2^i x^2+a_1^i x+a_0,
\end{equation}
where $[a_4^i, a_3^i, a_2^i, a_1^i, a_0^i]$ denote polynomial coefficients, and $i (i= 1, 2, 3)$ represents three principal lines of the palmprint. 
Compared to the second-order B\'ezier curves, a fourth-order polynomial representation provides sufficient expressiveness to capture the smoothness of palm creases.
Additionally, considering the physiological basis and distinct distribution of palm creases, it is essential to sample curve parameters that reflect the actual distribution.

As shown in Fig.\ref{fig:poly} (b), we manually label the points on the three principal lines of 1000 palmprint ROI images selected from publicly available datasets.
For each line, the polynomial coefficients are calculated using the least squares method, based on $n$ labeled points $((x_0,y_0 ),(x_1,y_1 ), \ldots, (x_{n-1},y_{n-1}))$, as follows,
\begin{equation}
    \mathbf a^\mathrm{T} =(\mathbf X^ \mathrm T \mathbf X)^{-1} \mathbf X^ \mathrm T \mathbf y,    
\end{equation}
where $\mathbf a^\mathrm{T}$ represents coefficient vector $[a_0,a_1,\ldots,a_{4}]^ \mathrm T$, $\mathbf y$ represents vector $[y_0,y_1,\ldots,y_{n-1}]^\mathrm{T}$, and $\mathbf X \in \mathbb R^{n \times 5}$ is a Vandermonde matrix. 
We apply this formula to get coefficient vectors $(\mathbf a^1, \mathbf a^2, \mathbf a^3)$ of three principal lines.

\begin{figure}[!tb]
   \centering
   \center{\includegraphics[width=.95\linewidth]  {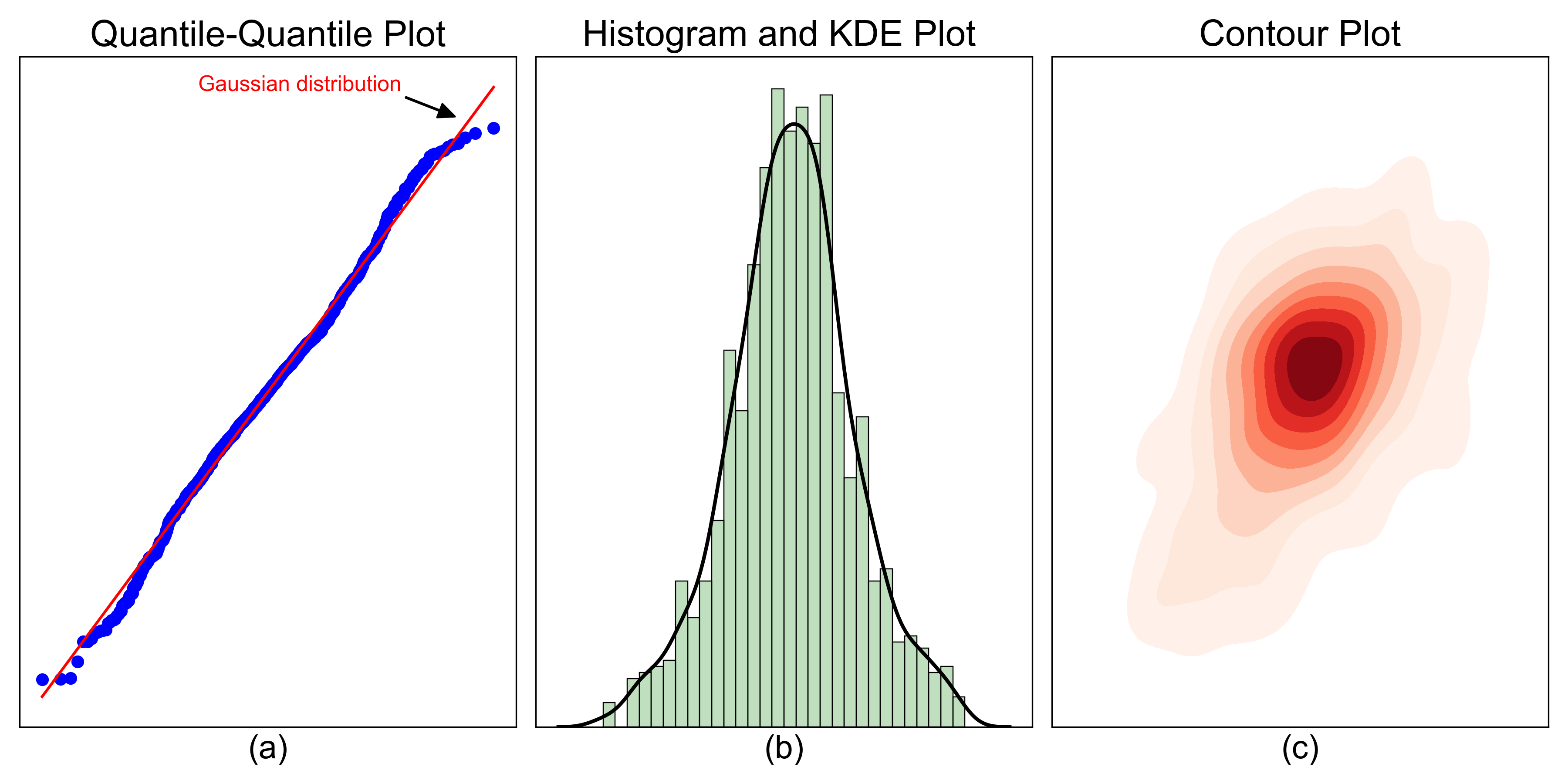}}
   \caption{Statistical analysis of polynomial coefficients: (a) Quantile-Quantile Plot of coefficients $a_4^1$. (b) Histogram and Kernel Density Estimate (KDE) Plot of coefficients $a_4^1$. (c) Contour Plot of coefficients $a_4^1$ and $a_3^1$.
   }\label{fig:hist}
   \vspace{-3mm}
\end{figure}

In this way, polynomial coefficients of three principal lines $(\mathbf a^1,\mathbf a^2,\mathbf a^3)$ are calculated 
for each palmprint sample. 
Then, we conduct a statistical analysis of the polynomial coefficients. 
Partial visualization results are shown in Fig.\ref{fig:hist}, from which we observe that the coefficient approximately follows a Gaussian distribution.
More statistical results are provided in the supplementary materials.
Consequently, we represent the distribution of $a_j^i$ as $p(a_j^i) \sim N(\mu_j^i,(\sigma^2)_j^i)$, where $\mu_j^i$ and $(\sigma^2)_j^i$ denote mean and variance 
of $a_j^i$. 
Therefore, the coefficient vector $\mathbf a^i$  follows the joint distribution $p(a_0^i,a_1^i,\ldots, a_4^i)$, which is a multivariate Gaussian distribution expressed as follows,
\begin{equation}
     p(\mathbf a^i) \sim \mathcal N(\mathbf \mu^i, \Sigma^i), 
\end{equation}
where $\mathbf \mu^i$, $ \Sigma^i$ are the mean vector and covariance matrix of $\mathbf a^i$.
By employing the proposed representation and estimated distribution, the complex palm creases located on a plane are mapped into a parameter space that conforms to a multivariate Gaussian distribution. 
This allows us to effortlessly sample the polynomial coefficients from the statistical distribution and then produce three principal lines using polynomial representation. 

Additionally, we also record the coordinates of two endpoints of each line, i.e., the starting point  $x_s^i$ and the ending point $x_e^i$, and calculate their statistical distribution.
By following the same approach, we determine the range for synthesizing polynomial creases by sampling from the estimated distribution.
With the addition of minor random straight lines to simulate wrinkles, a pseudo palm crease image is generated, as shown in Fig.\ref{fig:poly}. 

\textbf{A palm creases similarity control mechanism} is proposed by scaling the variance of the estimated Gaussian distribution by a factor of $\gamma^2$.
Specifically, as $\mathbf a^i$ is sampled from $\mathcal N(\mathbf \mu^i, \gamma^2 \Sigma^i)$, a smaller $\gamma$ (less than 1) results in a higher degree of similarity in the generated palm creases, while a larger $\gamma$ (greater than 1) leads to lower similarity.
\subsection{Palm Creases conditioned Diffusion Model}
\begin{figure*}[!tb]
   \centering
   \center{\includegraphics[width=.95\linewidth]  {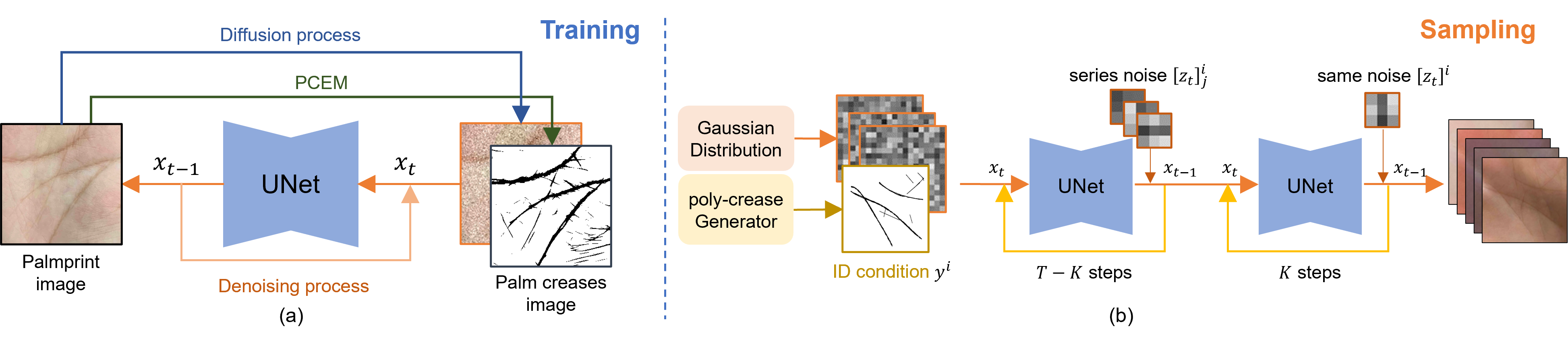}}
   \vspace{-3mm}
   \caption{The proposed intra-class variation controllable diffusion model. (a) Training process: palm crease images, extracted from palmprints using PCEM \cite{jin2024pce}, are employed as conditions and concatenated with diffused palmprint images, serving as input for the UNet. (b) Sampling process: polynomial creases, as synthetic identity, are first generated and adopted to create consistent samples. The $K$-step noise-sharing sampling is applied to obtain palmprint datasets with varying degrees of intra-class variations.
   }\label{fig:diffusion}
   \vspace{-3mm}
\end{figure*}
\begin{figure}[!tb]
   \centering
   \center{\includegraphics[width=\linewidth]  {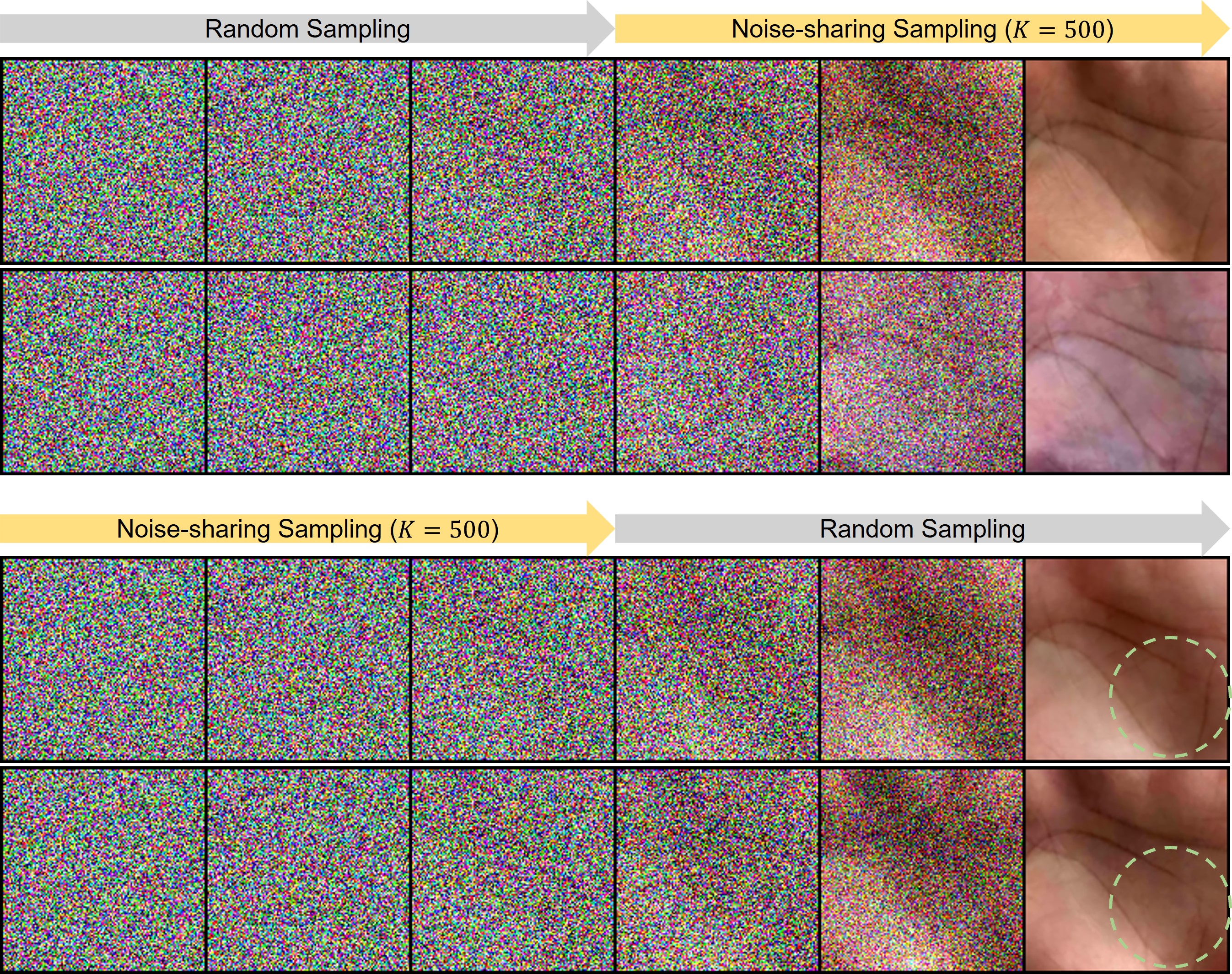}}
   \caption{Generated palmprint results under different noise-sharing strategies. The figure illustrates the outcomes of applying noise-sharing in the last $K=500$ steps (top) versus the first $K=500$ steps (bottom) of a total $T=1000$ steps during the sampling process for the same identity.
   }\label{fig:steps}
   \vspace{-3mm}
\end{figure}

Diffusion models \cite{croitoru2023diffusion} are trained to recover an image from random noise. In contrast to GANs, which consist of two networks and utilize an adversarial loss, diffusion models adopt two processes: the diffusion process and the denoising process.
The diffusion process is typically defined as a Markov chain composed of several diffusion steps. 
At each step, the model introduces a small amount of Gaussian noise with varying variances into the data.
After a total of $T$ steps, the data is degraded to pure Gaussian noise. 
The diffusion process \cite{croitoru2023diffusion} is expressed as,
\begin{equation}
p(x_t | x_{t-1}) = \mathcal N (x_t; \sqrt{1-\beta_t} x_{t-1}, \beta_t I),    
\end{equation}
where $x_t$ and $x_{t-1}$ represent the data at diffusion step $t$ and $t-1$ respectively, and $\beta_t$ is the variance schedule determined in advance that controls the amount of noise added.

The denoising process aims to learn the reverse of the diffusion process, and recover data $x_0$ from Gaussian noise $x_T$ step by step. 
An UNet network is trained to predict the noise at the $t$ step and thereby restore $x_{t-1}$ from $x_t$.
Since we aim to use the diffusion model to generate palmprint datasets, 
unconditional diffusion is insufficient to meet our requirements. 
To control the identity of generated palmprint images, we use the palm crease image as a condition. 
The consistent palm crease image is extracted from the real palmprint image using a palm creases extraction module (PCEM) \cite{jin2024pce}.
Then, the palm creases are adopted to govern the identity of the generated palmprints.
Therefore, we choose to integrate the condition image into the UNet through channel concatenation.
Specifically, paired palm crease images and corrupted palmprint images are concatenated and sent to the network, as shown in Fig.\ref{fig:diffusion}(a). We use the following training objective \cite{croitoru2023diffusion},
\begin{equation}
     L = E_{t, x_t, \epsilon} \left[ \| \epsilon - \epsilon_\theta\left(x_t, t, y\right) \|_2^2 \right],     
\end{equation}
where $\epsilon_\theta$ represents the parameterized network, and $y$ stands for palm crease image.

\subsection{The \textit{K}-step Noise-sharing Sampling}
To generate new samples, the diffusion model follows the reverse diffusion process.
Starting from random Gaussian noise, the model iteratively applies the learned denoiser to remove noise and generate final outputs.
The iterative sampling process \cite{croitoru2023diffusion} is expressed as follows,
\begin{equation}
     x_{t-1} = \frac{1}{\sqrt{\alpha_t}}\left(  x_t - \frac{1-\alpha_t}{\sqrt{1 - \bar{\alpha_t}}} \epsilon_\theta (x_t, t, y)  \right) + \sigma_t z_t, \label{sampling} 
\end{equation}
where $\alpha_t$ is defined as $1-\beta_t$, and $\bar{\alpha_t} := \Pi_{i=0}^t \alpha_i$. The $\sigma_t$ is step-dependent constants, and $z_t$ is a random noise sampled from $\mathcal N (\mathbf{0}, \mathbf{I})$. 
After the iterative process concludes, the estimate of $x_0$, derived from the final iteration, is the synthesized palmprint.
To control the diversity of generated results, many diffusion-based methods adopt text prompts \cite{croitoru2023diffusion, grosz2024genpalm}. 
However, palmprint images are relatively simple, comprising primarily of crease and skin components, which complicates the use of text descriptions to govern diversity. 
We have observed that the random noise added at each step provides the randomness of the sampling process. 
Therefore, by controlling the random noise added to samples within the same identity, we manipulate the intra-class variation of the generated datasets effectively.

The $K$-step noise-sharing sampling is proposed to generate palmprint datasets with varying degrees of intra-class variations.
Specifically, suppose generating the $j$-th sample of the $i$-th identity with the palm crease image $y^i$, we rewrite sampling process Eq.\ref{sampling} as follows:
\begin{equation}
     \textstyle [x_{t-1}]_j^i = \frac{1}{\sqrt{\alpha_t}}\left(  [x_t]_j^i - \frac{1-\alpha_t}{\sqrt{1 - \bar{\alpha_t}}} \epsilon_\theta ([x_t]_j^i, t, y^i)  \right) + \sigma_t [z_t]_j^i,     
\end{equation}
where $[z_t ]_j^i$ is the random noise sampled from a standard Gaussian distribution, influencing the diversity of samples.
We aim to utilize the same shared noise $[z_t ]^i$, instead of a series of different $[z_t ]_j^i$ to samples under the same identity condition. 
We select a continuous $K$-step sequence within the total $T$ steps and apply this, as shown in Fig.\ref{fig:diffusion}(b).

Additionally, we have found that the application of noise-sharing in the first $K$ steps and the last $K$ steps during the sampling process exhibits distinct behaviors. 
With $K=500$ and a total of $T=1000$ steps, the sampling process under the same identity yields the results shown in Fig.\ref{fig:steps}. 
When noise-sharing is applied in the first $K$ steps, the generated results maintain a consistent style, while the details of the palm creases exhibit minor variations, leading to a generated dataset with low identity consistency. 
Conversely, when noise-sharing is applied in the last $K$ steps, the generated results present diverse styles, but the consistency of the palm crease details improves. 
This suggests that during the sampling process, the style information of the palmprint is generated first, followed by the restoration of texture details.

Therefore, we apply noise-sharing in the last $K$ steps. As $K$ increases, the crease consistency in the sampled results under the same identity tends to improve. However, when $K$ continues to increase, the stylistic diversity of the sampled results becomes restricted. 
We conduct experiments on datasets generated using different values of $K$.

\begin{figure*}[!tb]
   \centering
   \small
   \center{\includegraphics[width=.95\linewidth]  {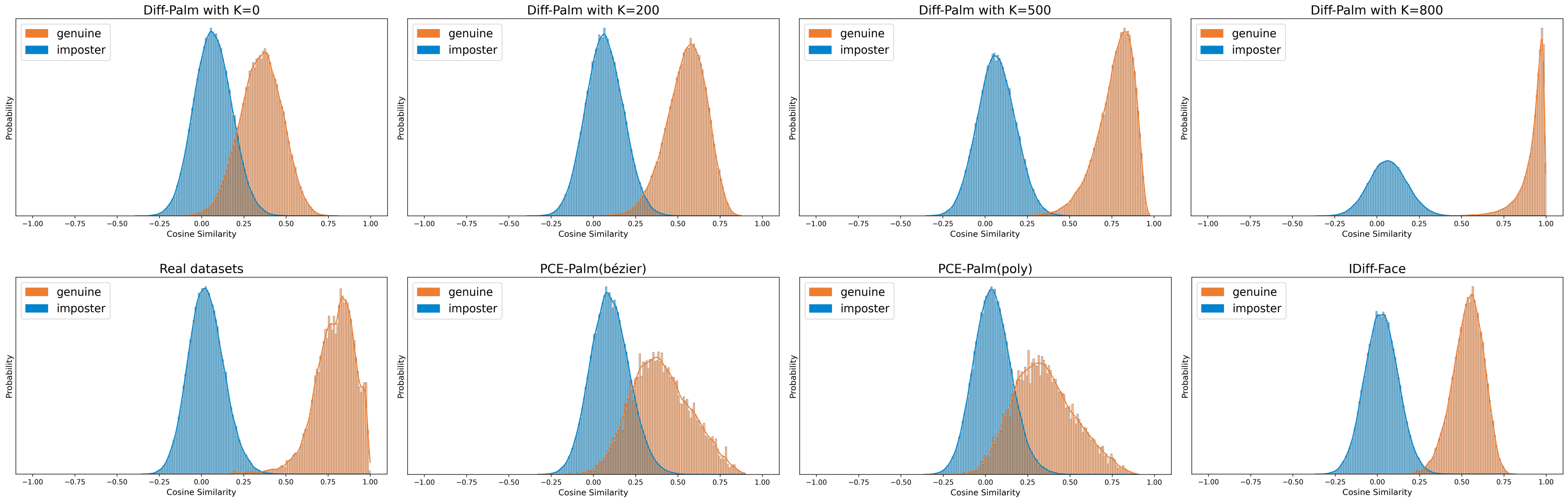}}
   \caption{Comparison score plots for synthetic datasets generated by different methods and real datasets. The genuine and imposter comparison scores are calculated with features, which are extracted by a pre-trained ArcFace model from datasets.
   }\label{fig:scores}
   \vspace{-3mm}
\end{figure*}

\begin{figure}[!tb]
   \centering
   \small
   \center{\includegraphics[width=.95\linewidth]  {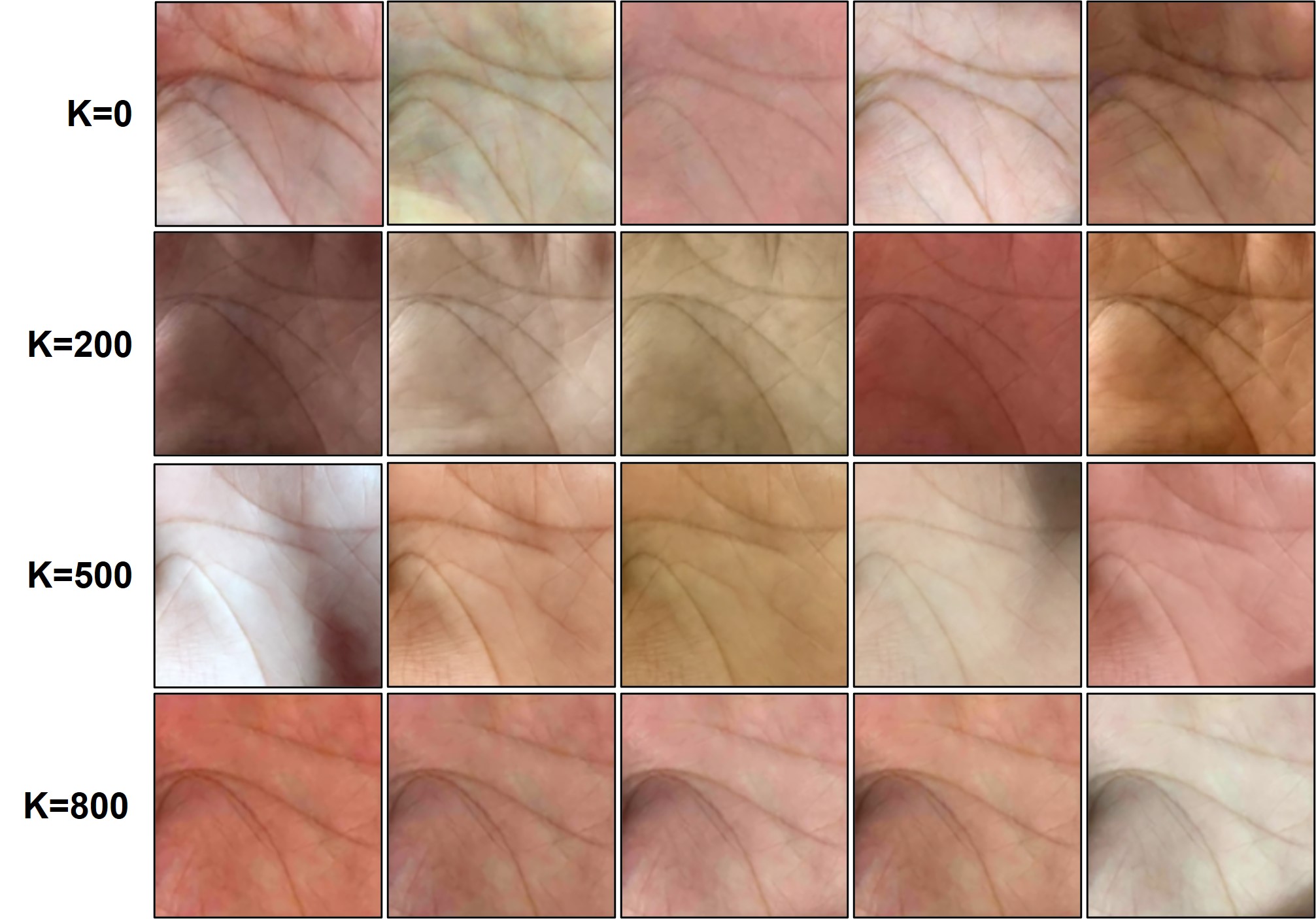}}

   \caption{Example images of datasets generated by Diff-Palm with $K$-step sharing-noise sampling for $K=0, 200, 500$, and $800$.
   }
   \label{fig:intra}
   \vspace{-3mm}
\end{figure}

\section{Experiments}
\subsection{Experimental Setups}
We conduct experiments under an open-set protocol, splitting each public dataset into training and test sets with a 1:1 ratio with no overlapping identities (IDs).
We utilize the TAR (True Accept Rate) @FAR (False Accept Rate) metric to evaluate the performance of recognition models.

\textbf{Datasets}
We do not use public datasets to train our generative models to ensure a fair comparison,
which also places our method in a more challenging cross-dataset setting.
Instead, we adopt an anonymous dataset collected from the Internet,
which has been processed to include 48,000 palmprint images.
We utilize publicly available datasets for evaluation and to train recognition models as baselines.
Experiments are conducted on 
following public datasets:  
CASIA \cite{sun2005ordinal}, 
PolyU \cite{zhang2003online}, 
Tongji \cite{zhang2017towards}, 
MPD \cite{zhang2020towards}, 
XJTU-UP \cite{shao2020effective}, 
IITD \cite{kumar2008incorporating},
and NTU-CP-v1 \cite{matkowski2019palmprint}.

\textbf{Generation Model Training Setups}
Our conditional diffusion models utilize a UNet backbone \cite{dhariwal2021diffusion} that takes a four-channel input and 64 base channels with five resolution levels. 
The multipliers for the number of channels used on those levels are $1$, $1$, $2$, $3$, and $4$, respectively.
Attention blocks are applied in residual blocks of the last resolution level. 
During the training process, we adopt the AdamW optimizer \cite{loshchilov2017decoupled}, set the learning rate $1e-4$, and train the network for a total of $30k$ steps. 
We apply an EMA to the weights of the model with an exponential factor of $0.9999$.
The batch size is $64$ and is equally split across $4$ GPUs. 
For the diffusion process, we set $T = 1000$ steps and adopt a linear diffusion variance schedule.
In subsequent experiments, we default to generating datasets with $2k$ IDs and 20 samples per ID.

\textbf{Recognition Model Training Setups}
We train ArcFace models \cite{deng2019arcface} with a modified Resnet-18 \cite{he2016deep} backbone for palmprint recognition. 
Palmprint images are resized to $112\times 112$ and trained with ArcFace loss ($m=0.5$, $s=64$) over 20 epochs using a stochastic gradient descent (SGD) optimizer. 
We use an initial learning rate of $1e-1$, momentum of $0.9$, weight decay of $5e-4$, and a step learning rate schedule, which divides the learning rate by $10$ at $7$-th and $15$-th epoch. 
RandAugment \cite{cubuk2020randaugment} is adopted for data augmentation with parameters ($4$, $4$).
We double the number of IDs in the training set by horizontally flipping the palmprint images.
Training is implemented on 4 NVIDIA V100 GPUs with a batch size of $256$. 
Consistent hyperparameters are used for all recognition models across various datasets.
\subsection{Experimental Results}

\begin{figure}[!tb]
   \centering
   \small
   \center{\includegraphics[width=.95\linewidth]  {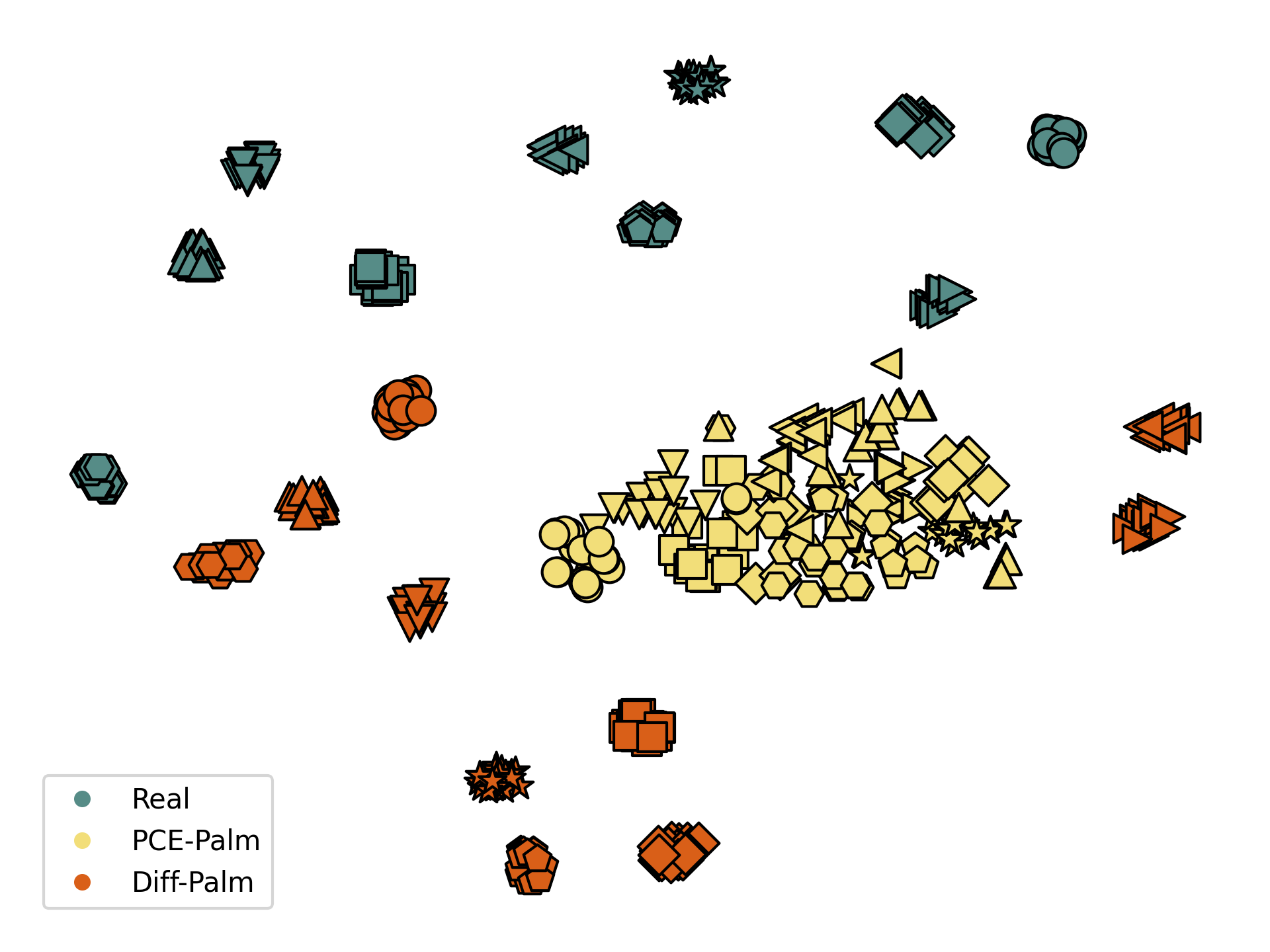}}
   \vspace{-3mm}
   \caption{The t-SNE polt of features extracted from real, PCE-Palm generated, and Diff-Palm generated datasets. Colors denote different datasets, while shapes within the same color correspond to different identities.
 }\label{fig:tsne}
 \vspace{-3mm}
\end{figure}

\begin{table*}[tb]
    \centering
    \small
    
    \begin{tabular}{lcccccccccccc}
    \toprule
    \multirow{3}{*}{\textbf{Methods}} & \multicolumn{2}{c}{\textbf{Configs}} & 
    \multicolumn{4}{c}{\textbf{ Score Distributions}} &
    \multicolumn{6}{c}{\textbf{Performance (TAR@FAR=1e-6) $\uparrow$}} \\
    \cmidrule(lr){2-3} \cmidrule(lr){4-7} \cmidrule(lr){8-13}  
    & \multirow{2}{*}{$K$} & \multirow{2}{*}{$\gamma$} &
    \multicolumn{2}{c}{genuine} & \multicolumn{2}{c}{imposter} & 
    \multirow{2}{*}{CASIA} & \multirow{2}{*}{PolyU} &  \multirow{2}{*}{TongJi} &  
    \multirow{2}{*}{MPD} & \multirow{2}{*}{XJTU-UP} &  \multirow{2}{*}{\textit{Avg.}} \\
    & & & mean &  std &  mean &  std & & & & & & \\
    \midrule
    \foretab Diff-Palm & 0 & 1.0     & 0.350 & 0.124 & 0.069 & 0.112    & 0.7278 & 0.7392 & 0.5151 & 0.3406 & 0.1527 & 0.4951 \\
    \foretab Diff-Palm & 200 & 1.0   & 0.549 & 0.118 & 0.069 & 0.111    & 0.8420 & 0.9111 & 0.8113 & 0.4237 & 0.6018 & 0.7180 \\
    \foretab Diff-Palm & 500 & 1.0   & 0.766 & 0.108 & 0.068 & 0.112    & \textbf{0.8782} & \textbf{0.9601} & \textbf{0.9460} & \textbf{0.4643} & \textbf{0.6161} & \textbf{0.7729} \\
    \foretab Diff-Palm & 800 & 1.0   & 0.922 & 0.072 & 0.069 & 0.112    & 0.8354 & 0.9014 & 0.8592 & 0.3369 & 0.5035 & 0.6873 \\
    \midrule
    PCE-Palm(B\'ezier) & --  & --  & 0.402 & 0.171 & 0.099 & 0.116      & 0.6796 & 0.7740 & 0.6149 & 0.2972 & 0.4150 & 0.5561 \\
    PCE-Palm(poly)   & -- & 1.0  & 0.351 & 0.186 & 0.041 & 0.110      & 0.8220 & 0.8687 & 0.7674 & 0.4373 & 0.5974 & 0.6986 \\
    Diff-Palm(B\'ezier)& 500 & --  & 0.629 & 0.198 & 0.171 & 0.114      & 0.6722 & 0.8236 & 0.7246 & 0.2732 & 0.3658 & 0.5719 \\
    Diff-Palm(poly)  & 500 & 1.0 & 0.766 & 0.108 & 0.068 & 0.112      & \textbf{0.8782} & \textbf{0.9601} & \textbf{0.9460} & \textbf{0.4643} & \textbf{0.6161} & \textbf{0.7729} \\ 
    \midrule
    \foretab Diff-Palm & 500 & 0.25  & 0.748 & 0.121 & 0.107 & 0.113  & 0.8440 & 0.9134 & 0.8410 & 0.4162 & 0.5794 & 0.7188 \\
    \foretab Diff-Palm & 500 & 0.5   & 0.764 & 0.118 & 0.088 & 0.117     & 0.8613 & 0.9230 & 0.8700 & 0.4268 & 0.5823 & 0.7327 \\
    \foretab Diff-Palm & 500 & 1.0   & 0.766 & 0.108 & 0.068 & 0.112    & \textbf{0.8782} & \textbf{0.9601} & \textbf{0.9460} & \textbf{0.4643} & \textbf{0.6161} & \textbf{0.7729} \\
    \foretab Diff-Palm & 500 & 2.0 & 0.784 & 0.102 & 0.070 & 0.109    & 0.8053 & 0.9308 & 0.9162 & 0.3697 & 0.4890 & 0.7022 \\    
    \foretab Diff-Palm & 500 & 4.0 & 0.780 & 0.103 & 0.095 & 0.108    & 0.7793 & 0.8804 & 0.8662 & 0.3041 & 0.4341 & 0.6528 \\ 
    \midrule
    Diff-Palm(first $K$) & 500 & 1.0 & 0.475 & 0.043 & 0.117 & 0.11  & 0.7649 & {0.7571}  & {0.5136} &{0.2502} & 0.0307 & {0.4633} \\
    Diff-Palm(last $K$) & 500 & 1.0 & 0.766 & 0.108 & 0.068 & 0.112  & \textbf{0.8782} & \textbf{0.9601}  & \textbf{0.9460} & \textbf{0.4643} & \textbf{0.6161} & \textbf{0.7729} \\
    \bottomrule
    \end{tabular}

    \caption{
    Verification performance of recognition models(ArcFace \cite{deng2019arcface}) 
    with the same hyperparameters used for various generated datasets, and without any fine-tuning on real data. 
    Synthetic datasets default to be generated with $2k$ IDs and 20 samples per ID. 
 }
    \label{tab:compare}
    \vspace{-3mm}
\end{table*}

\begin{table}[!tb]
    \centering
    \small
    \resizebox{1\linewidth}{!}{
    \begin{tabular}{lccccc}
    \toprule
    Methods         &$K$& $C_{identity}$  & $D_{intra}$ & $U_{class}$  & Perf.\\
    \midrule
    \foretab Diff-Palm       & 0     & 0.5923    & 1.000     & 0.897 & 0.4951 \\
    \foretab Diff-Palm       & 200   & 0.9667    & 0.9615    & 0.980 & 0.7189\\
    \foretab Diff-Palm       & 500   & 0.9981    & 0.9372    & 0.996 &\textbf{0.7729} \\
    \foretab Diff-Palm       & 800   & 0.9996    & 0.6659    & 0.999 & 0.6873 \\
    \midrule    
    PCE-Palm(B\'ezier)   & --    & 0.6046    & 1.1894    & 0.653 & 0.5561 \\
    PCE-Palm(poly)       & --    & 0.6850    & 1.1816    & 0.958 & 0.6986 \\
    Diff-Palm(B\'ezier)  & 500   & 0.9952    & 0.9691    & 0.694 & {0.5719} \\
    Diff-Palm(poly)      & 500   & 0.9981    & 0.9372    & 0.996 &\textbf{0.7729} \\
    \bottomrule
    \end{tabular}}
    \caption{Evaluation of synthetic datasets. $C_{identity}$, $D_{intra}$, and $U_{class}$ represent identity consistency,  intra-class diversity, and class uniqueness, respectively.}
    \label{tab:eval}
    \vspace{-3mm}
    
\end{table}

\textbf{Evaluating Intra-Class Variation}
We generate datasets by Diff-Palm with $K$-step sharing-noise sampling for $K=0, 200, 500$ and $800$. 
Then, we use the ArcFace model pre-trained on real datasets to extract features from synthesized datasets and calculate genuine and imposter score distributions. 
Comparison score plots and example images are shown in Fig.\ref{fig:scores} and Fig.\ref{fig:intra} respectively. 
As $K$ increases, intra-class variation decreases, and the genuine distribution shifts right. 
The verification performance in Tab.\ref{tab:compare} reveals that the recognition model trained on the synthesized dataset with $K=500$ achieves higher recognition performance.

\textbf{Comparing Palm Crease Representations}
We compare the performance of PCE-Palm \cite{jin2024pce} and our diffusion model by training them on 
different types of palm crease representations:
B\'ezier creases \cite{zhao2022bezierpalm} and polynomial creases, respectively.
Each model is trained using consistent hyperparameters, with the only difference being the type of crease images used.
As evidenced in Tab.\ref{tab:compare}, the polynomial creases method significantly outperforms the B\'ezier creases. 
This result indicates that our polynomial crease representation more accurately bridges the gap between synthetic and real palmprints.

\textbf{Evaluating Identity and Diversity Metrics}
We adopt identity consistency, intra-class diversity, and class uniqueness metrics from \cite{kim2023dcface} to evaluate the respective properties of synthetic datasets.
As shown in Tab.\ref{tab:eval}, for our Diff-Palm, there exists a trade-off between identity consistency and intra-class diversity as parameter $K$ is adjusted. 
For $K=500$, the generated dataset strikes a balance between them, leading to optimal recognition performance. 
When comparing generative methods with different palm crease representations, it is evident that the use of B\'ezier creases leads to a noticeable decline in class uniqueness, as opposed to polynomial creases.
Our analyses indicate that the superior performance of our method is due to the \textit{effective balance between identity consistency and intra-class diversity} achieved by the intra-class variation controllable method, as well as the \textit{enhanced class uniqueness} resulting from the use of polynomial creases.
In contrast, PCE-Palm suffers from low identity consistency and reduced class uniqueness, leading the poor performance.
These analysis results are also validated by the t-SNE visualization, as shown in Fig.\ref{fig:tsne}.
Further Details about these metrics are provided in the supplementary materials.

\begin{table*}[!tb]
    \centering
    \small
    
    \begin{tabular}{lcccccccccc}
    \toprule
    \multirow{2}{*}{\textbf{Methods}} & \multicolumn{4}{c}{{\textbf{Configs}}} & \multicolumn{6}{c}{{\textbf{Performance (TAR@FAR=1e-6) $\uparrow$}}} \\
    \cmidrule(lr){2-5} \cmidrule(lr){6-11} 
    & \multirow{1}{*}{\#IDs} & \multirow{1}{*}{\# per ID} & \multirow{1}{*}{\#Images} & \multirow{1}{*}{FT w/ Real} &
    \multicolumn{1}{c}{CASIA} & \multicolumn{1}{c}{PolyU} & \multicolumn{1}{c}{TongJi} &
    \multicolumn{1}{c}{MPD} & \multicolumn{1}{c}{XJTU-UP} & \multicolumn{1}{c}{\textit{Avg.}} \\
    \midrule
    \foretab Real data& 2.2k & -- & 29.3k & \XSolidBrush & \textbf{0.9200} & 0.9196 & 0.9209 & 0.3877 & \textbf{0.6247} & 0.7546 \\
    \foretab IDiff-Face \cite{boutros2023idiff} & 2k &20 & 40k  & \XSolidBrush & 0.7977 & 0.7720 & 0.6115 & 0.1847 & 0.2667 & 0.5265 \\
    \foretab Vec2Face \cite{wu2024vec2face} & 2k &20 & 40k  & \XSolidBrush & 0.8201    & 0.8105 &    0.8199  & 0.2575 & 0.2578 & 0.5932\\
    \foretab PCE-Palm \cite{jin2024pce} & 2k &20 & 40k    & \XSolidBrush & 0.6796 & 0.7740 & 0.6149 & 0.2972 & 0.4150 & 0.5561 \\
    \foretab Diff-Palm & 2k &20 & 40k        & \XSolidBrush & 0.8782 & \textbf{0.9601} & \textbf{0.9460} & \textbf{0.4643} & 0.6161 & \textbf{0.7729} \\
    \midrule
    PCE-Palm \cite{jin2024pce} & 2k &20 & 40k    & \Checkmark & 0.9415 & 0.9792 & 0.9393 & \textbf{0.6662} & 0.7948 & 0.8642 \\
    Diff-Palm & 2k &20 & 40k        & \Checkmark & \textbf{0.9787} & \textbf{0.9859} & \textbf{0.9744} & 0.5979 & \textbf{0.8433} & \textbf{0.8760} \\
    \midrule 
    \foretab Diff-Palm & 5k &20 & 100k  & \XSolidBrush & 0.8972 & 0.9757 & 0.9590 & 0.5048 & 0.7534 & 0.8180 \\
    \foretab Diff-Palm & 10k &20 & 200k  & \XSolidBrush & 0.9072 & 0.9818 & 0.9632 & 0.5687 & 0.8143 & 0.8470 \\
    \foretab Diff-Palm & 20k &20 & 400k  & \XSolidBrush & 0.9383 & 0.9795 & 0.9783 & 0.6304 & 0.8655 & 0.8784 \\
    \foretab Diff-Palm & 30k &20 & 600k  & \XSolidBrush & 0.9472 & 0.9772 & 0.9772 & 0.6651 & 0.9015 & 0.8937 \\
    \foretab Diff-Palm & 40k &20 & 800k  & \XSolidBrush & 0.9450 & 0.9799 & 0.9802 & 0.6641 & 0.9043 & 0.8947 \\ 
    \foretab Diff-Palm & 50k &20 & 1M    & \XSolidBrush & 0.9542 & 0.9822 & \textbf{0.9848} & 0.6814 & \textbf{0.9032} & 0.9011 \\
    \foretab Diff-Palm & 60k &20 & 1.2M    & \XSolidBrush & \textbf{0.9557} & \textbf{0.9865} & 0.9825 & \textbf{0.6856} & 0.9029 & \textbf{0.9026} \\
    \midrule
    Diff-Palm & 2k &50 & 100k & \XSolidBrush & 0.8810 & {0.9335} & {0.9185} &{0.4779} & 0.7214 & {0.7865} \\
    Diff-Palm & 2k &100 & 200k & \XSolidBrush & 0.9006 & {0.9436} & {0.9443} &{0.5522} & 0.7482 & {0.8178} \\
    Diff-Palm & 2k &150 & 300k & \XSolidBrush & 0.8817 & {0.9669} & {0.9482} &{0.5559} & 0.8043 & {0.8314} \\
    Diff-Palm & 2k &200 & 400k & \XSolidBrush & \textbf{0.9143} & \textbf{0.9710} & \textbf{0.9455} & \textbf{0.5775} & \textbf{0.8494} & \textbf{0.8515} \\
    \bottomrule
    
    \end{tabular}
    \vspace{-2mm}
    \caption{Verification performance of the recognition models across five public datasets, with the same hyperparameters used for both real and synthesized datasets. Diff-Palm is trained with $K=500$ and $\gamma=1.0$ by default. 
    }
    \label{tab:final}
    \vspace{-3mm}
    
\end{table*}

\textbf{Effect of Palm Crease Similarity on Recognition}
We synthesize polynomial crease datasets with varying $\gamma$ values to generate palmprint datasets through diffusion models. 
As shown in  Tab.\ref{tab:compare}, results indicate that recognition performance declines when $\gamma$ is either greater than 1 or less than 1, corresponding to the expansion or contraction of the estimated distribution's variance. 
We believe that when $\gamma > 1$, the generated data disrupts original distribution patterns, 
while when $\gamma < 1$ , training recognition models becomes increasingly difficult. 
These results confirm the reasonableness and validity of our estimated distributions.

\textbf{Comparing with Real Datasets and Other Methods}
We utilize seven public datasets to create a mixed dataset containing 2.2k IDs.
The current SOTA palmprint generation method, PCE-Palm \cite{jin2024pce}, 
and the face generation method \cite{boutros2023idiff, wu2024vec2face}, are trained for comparison. 
By default, we train the recognition models on synthetic datasets without employing real data fine-tuning. 
As shown in Tab.\ref{tab:final}, the recognition model trained on the dataset generated by our method outperforms the models trained on real data across three public datasets, and achieves an average improvement.
Furthermore, our approach significantly surpasses both PCE-Palm and facial models.
We also conduct fine-tuning experiments using real data for a comprehensive comparison with PCE-Palm.
More results are provided in the supplementary materials.

\textbf{Impact of Synthesized Identity and Sample Quantity}
We generate datasets with varying numbers of identities, maintaining $20$ samples per identity, to train recognition models.
As evidenced in Tab.\ref{tab:final}, the performance of the recognition model steadily improves as the number of identities increases.
The highest performance is observed when the number of identities reaches $60k$.
We also explore the impact of varying the number of samples per identity while keeping the number of identities constant.
As shown in Tab.\ref{tab:final}, increasing the number of samples per identity from $20$ to $200$ results in continuous performance improvement.
These findings demonstrate the scalability of our approach.

\textbf{Ablation and Discussion of $K$-Step Noise-Sharing Sampling}
We generate palmprint datasets using two different noise-sharing strategies (i.e., applied during the first $K$ steps or the last $K$ steps) and evaluate their performance. 
As shown in Tab.\ref{tab:compare}, applying noise-sharing during the first K steps results in lower intra-class diversity in the generated datasets, leading to poorer recognition performance. 
Additionally, the $K$ step noise-sharing sampling can serve as a plug-and-play method applicable to other diffusion-based generative models.
We provide results applying this method to IDiff-Face \cite{boutros2023idiff} in the supplementary materials.

\section{Conclusion}
This paper proposes a polynomial-based palm crease representation, capable of generating a large-scale dataset with a distribution closely resembling that of real palm crease. 
Subsequently, an intra-class variation controllable diffusion model is introduced, with a simple yet effective $K$-step noise-sharing sampling, which enables the generation of palmprint datasets with adjustable intra-class variations.
Experimental results indicate that our proposed method significantly reduces the gap between generated and real datasets. 
It is also the first time that the palmprint recognition model trained solely on our generated data, yet outperforms the model trained on real data.

\textbf{Acknowledgments}
We would like to acknowledge Zhiyuan Wang and Yang Wu for their assistance in polynomial creases.
This work is partly supported by the grants of the National Natural Science Foundation of China under Nos.62476077, 62076086, and 62272142.

{
    \small
    \bibliographystyle{ieeenat_fullname}
    \bibliography{main}
}

\clearpage
\setcounter{page}{1}
\maketitlesupplementary

This supplementary materials provide the following contents:
\begin{itemize}
    \item an overview of datasets used: both public datasets and anonymous datasets.
    \item evaluation metrics.
    \item additional experimental results.
    \item discussions on proposed sampling methods and validation approaches.
\end{itemize}

\section{Datasets}
\subsection{Public Datasets}
\begin{figure}[!tb]
   \centering
   \small
   \center{\includegraphics[width=.95\linewidth] {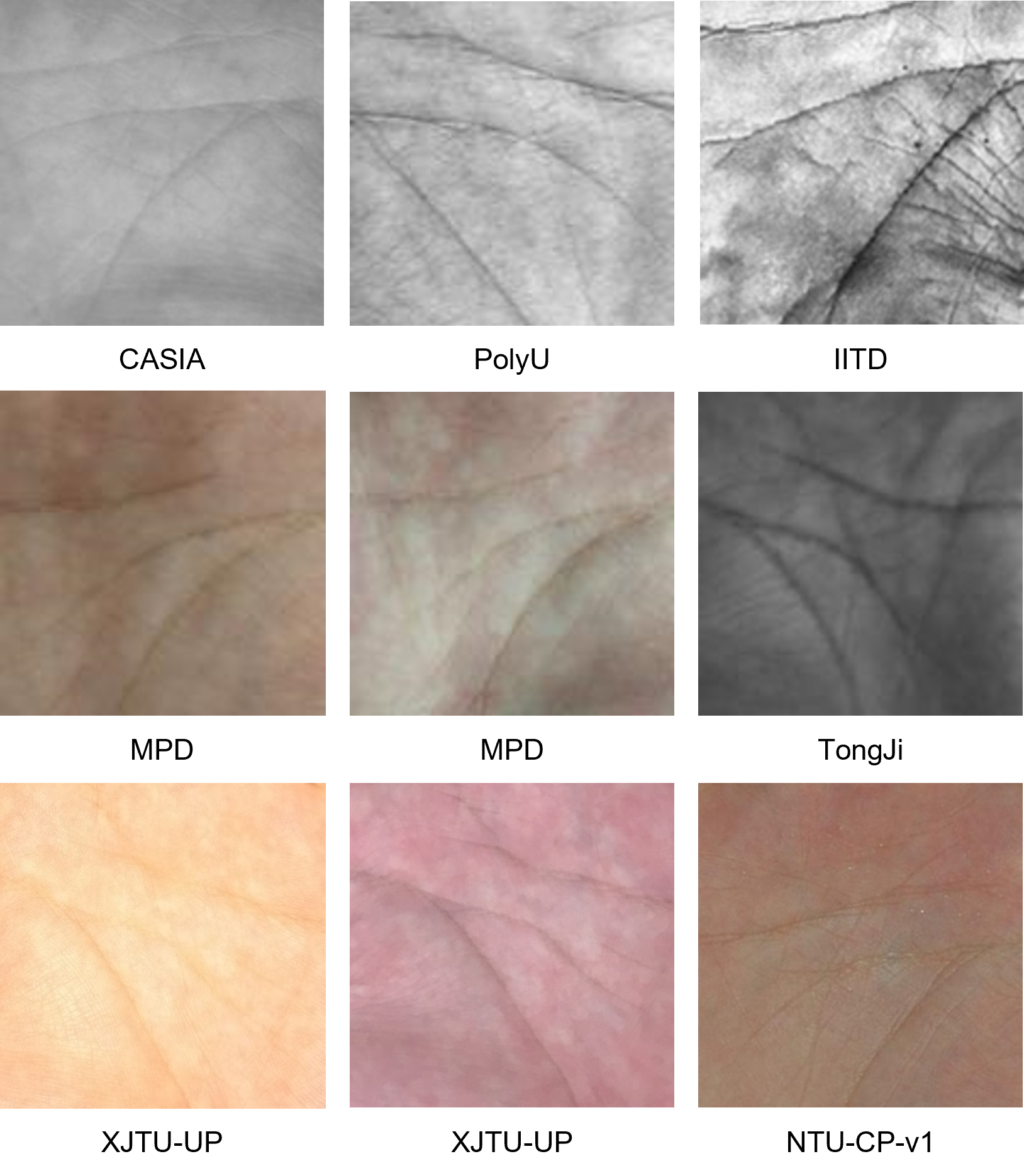}}
   \caption{Example images of public palmprint datasets.}
   \label{fig:datasets}
\end{figure}

We utilize seven publicly available palmprint datasets, 
including, CASIA \cite{sun2005ordinal}, 
PolyU \cite{zhang2003online}, 
Tongji \cite{zhang2017towards}, 
MPD \cite{zhang2020towards}, 
XJTU-UP \cite{shao2020effective}, 
IITD \cite{kumar2008incorporating},
and NTU-CP-v1 \cite{matkowski2019palmprint}, 
with detailed information provided in Tab.\ref{tab:datasets} and example images shown in Fig.\ref{fig:datasets}.
Following the open-set protocol, we divide the first five palmprint datasets into training and testing sets in a 1:1 ratio based on the number of IDs, 
ensuring no overlap between the IDs in the training and testing sets. 
Due to the limited number of images in the other two datasets, we used them exclusively for training.

\subsection{Collected Anonymous Dataset}
\begin{figure}[!tb]
   \centering
   \small
   \center{\includegraphics[width=.95\linewidth] {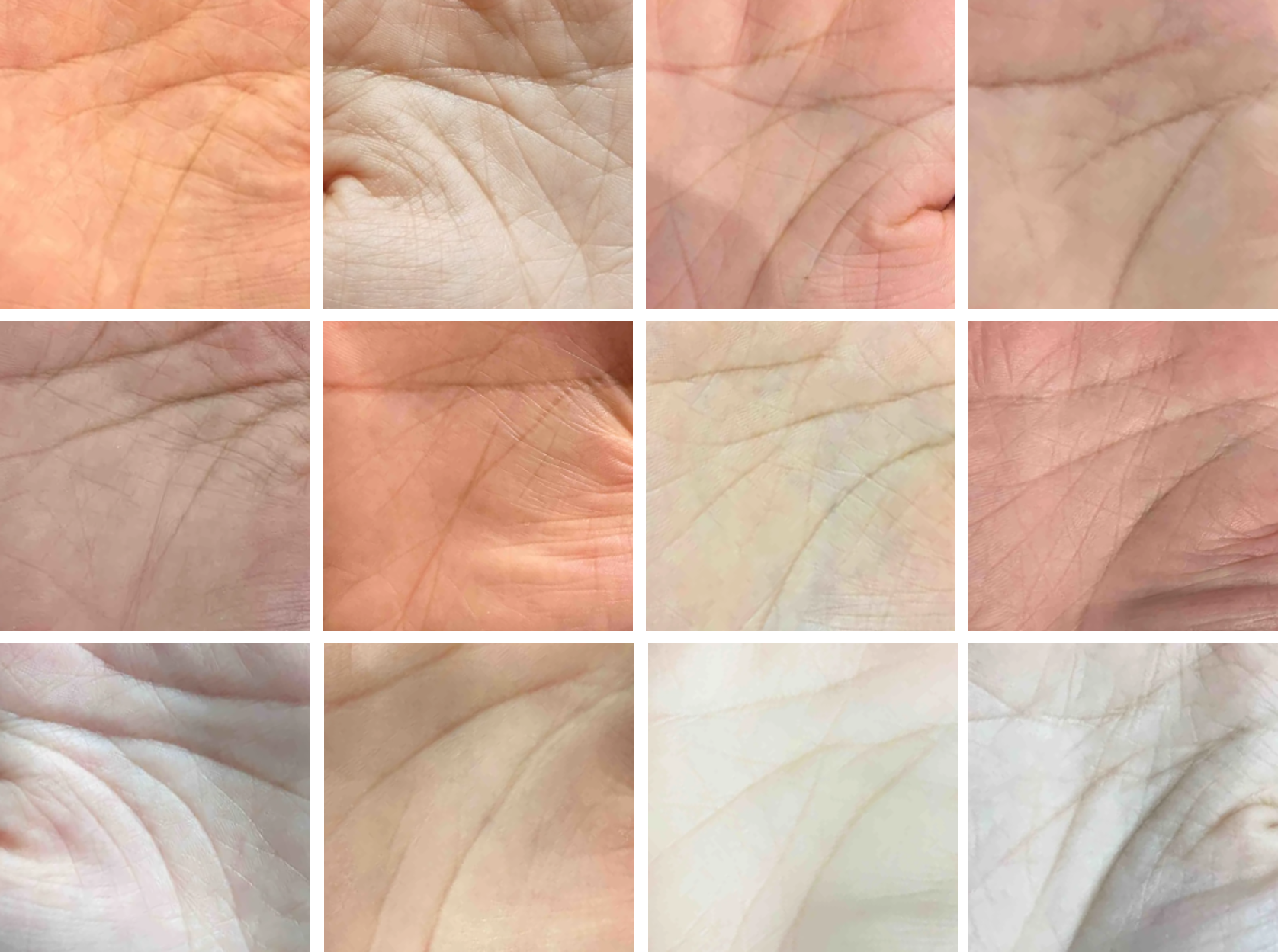}}
   \caption{Example images of collected anonymous datasets.}
   \label{fig:our-datasets}
\end{figure}

% \textbf{} 
We employ keywords such as "hand," "palm," and "palm print" to search for images on the Internet. 
After obtaining these images, we utilize Mediapipe \cite{lugaresi2019mediapipeframeworkbuildingperception} to detect the presence and completeness of palms in the images.  
Following this filtering process, we apply the detect-then-crop protocol in \cite{crop} to extract the Region of Interest (ROI) of the palmprints. 
We have acquired 48,000 complete palmprint ROI images, which are then used to train our generative model. 
% Additionally, we also experiment to verify whether there is any identity overlap between the collected anonymous dataset and publicly available datasets.
Example images are shown in Fig.\ref{fig:our-datasets}.
Due to relevant privacy protection regulations, we will release the URLs to these images.

\begin{table}[!b]
    \centering
    \small
    \begin{tabular}{lcccc}
    \toprule
    Datasets         & \#ID & \#Images & Devices  \\
    \midrule
    CASIA         & 620   & 5502    & Digital camera     \\
    PolyU         & 388   & 7738    & Scanner    \\
    Tongji        & 600   & 12,000    & Digital camera    \\
    MPD          & 400   & 16,000    & Mobile phone   \\
    XJTU-UP     & 200   & 7900    & Mobile phone    \\
    IITD        & 460   & 2601    & Digital camera    \\
    NTU-CP-v1     & 652   & 2390    & Digital camera  \\
    \bottomrule
    \end{tabular}
    \caption{Details of the seven public palmprint datasets.}
    \label{tab:datasets} 
\end{table}

\section{Evaluation Metrics}
\subsection{Performance Metrics}
we adopt the TAR(True accept ratio)@FAR(False accept ratio) metric, which is a widely used metric in open-set recognition tasks. 
It quantifies the system's ability to correctly accept genuine instances while controlling the rate of falsely accepted impostors. 
To compute TAR@FAR, one first determines the threshold $t$ under the specific FAR, 
which represents the proportion of non-genuine instances incorrectly accepted as genuine. 
At this threshold $t$, the TAR is calculated as the proportion of genuine instances correctly accepted, as follows,
$$
 \text{TAR}(t) = \frac{\text{Number of True Acceptances}}{\text{Total Number of Genuine Attempts}}.
$$
\begin{figure}[!tb]
   \centering
   \small
   \center{\includegraphics[width=.95\linewidth]  {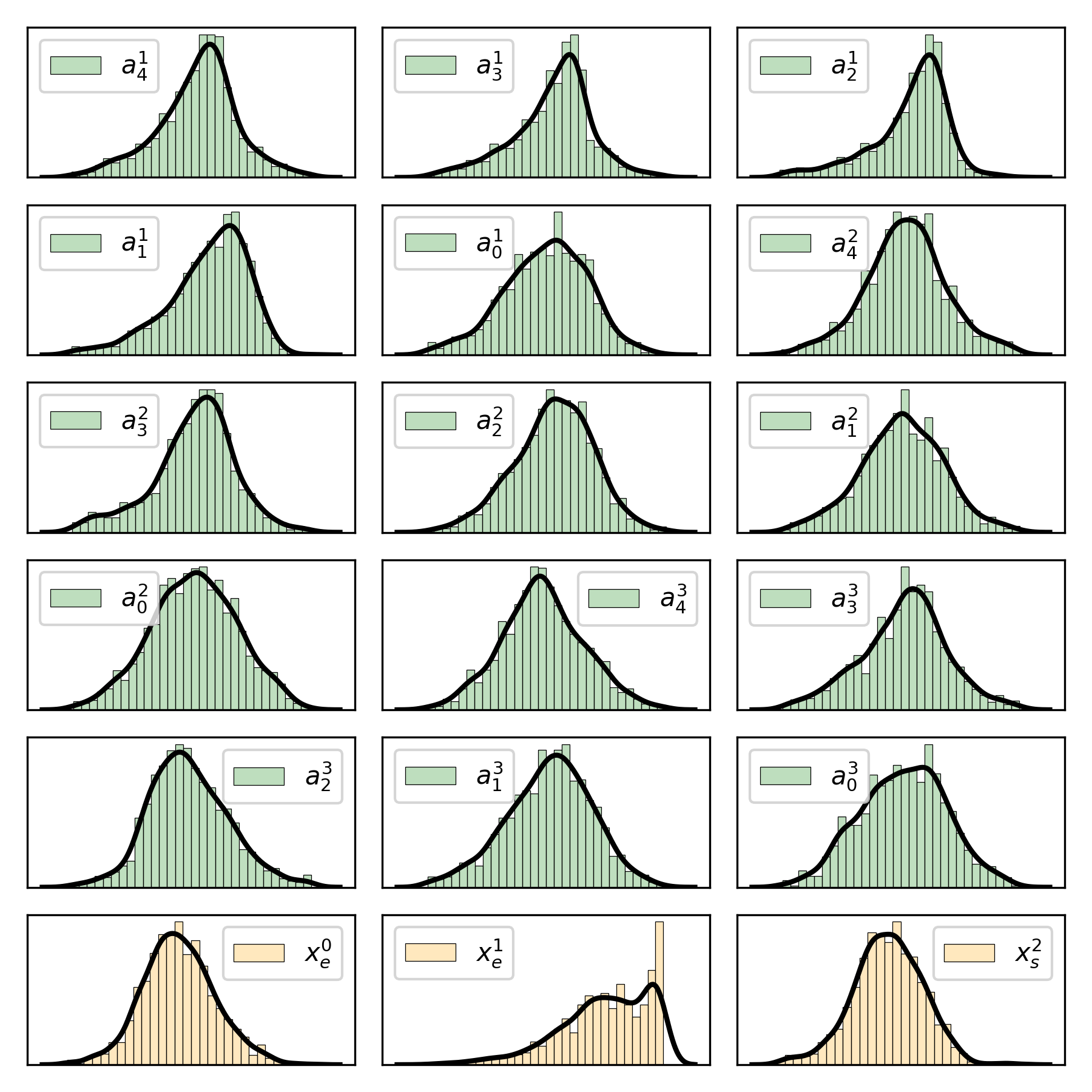}}
   \caption{Histograms of the 15 polynomial coefficients and the x-coordinates for 3 endpoints}\label{fig:poly-hist}
\end{figure}

\subsection{Datasets Evaluation Metrics}
We utilize three metrics derived from\cite{kim2023dcface} to evaluate the synthesized dataset. 
Specifically, we employ a recognition model $F_{eval}$, pre-trained on a real dataset to extract features from each image in the dataset. 
For the image $X_i^c$, $i$-th sample within $c$-th label, we denote its feature as $f^c_i = F_{eval}(X_i^c)$. 
We use cosine similarity to measure the distance between two samples. 
Additionally, we denote the feature of the center of each class as $\bar{f^c}$ for $c \in \{1, \dots C \}$, which is also the spherical mean of the samples within the same label.

\textit{Class Uniqueness.}
We first define $U_c$ as follows,
$$
U_c = \{ \bar{f^c}: d(\bar{f^{c_n}}, \bar{f^{c_m}}) > r, m < n, n, m \in \{1, \dots C\}  \},
$$
where $d(\cdot , \cdot)$ is the cosine distance. 
The $U_c$ is the set of unique subjects determined by the threshold $r$.
For this metric, we define $U_{class} := |U_c|/C$, the ratio between the number of unique subjects and the number of total labels.

\textit{Identity Consistency.} 
To measure how consistent the synthesized samples are in adhering to the label condition, 
we define $C_{identity}$ as
$$
C_{identity} = \frac{1}{C}\sum_{c=1}^{C}\frac{1}{N_c}\sum_{i=1}^{N_c} d(f_i^c, \bar{f^c}) < r,
$$
which is the ratio of individual features $f^c_i$ being close to the class center $f^c$. 
For a given threshold $r$, higher values of $C_{identity}$ mean the samples under the same label are more likely to be the same subject.

\textit{Intra-class Diversity.}
We aim to measure how diverse the generated samples are under the same label condition, as well as the diversity is in the style of an image, not in the subject’s identity.
In the original paper \cite{kim2023dcface}, the Inception Network pre-trained on ImageNet is utilized to extract the style information of the images. 
However, since palmprints are significantly different from the images in ImageNet, and are often simple and relatively uniform, we employ a pixel-based diversity measure. 
Specifically, we adopt $\bar{X^c} =\frac{1}{N_c} \sum_{i=0}^{N_c}X_i^c $ denotes the mean image of $c$ class, and diversity is defined as:
$$
D_{intra} = \frac{1}{C}\sum_{c=1}^{C}\frac{1}{N_c}\sum_{i=1}^{N_c} \| X_i^c - \bar{X^c} \|_1 ,
$$
where $\|\cdot\|_1$ denotes $L1$ norm.
We take the $D_{intra}$ value of datasets generated by Diff-Palm with $K=0$ as the baseline, normalizing it to $1.0$, and adjusting all other values accordingly.

\section{Additional Experimental Results}
% \begin{figure*}[!htb]
%    \centering
%    \small
%    \center{\includegraphics[width=.95\linewidth]  {./figures/all-scores.png}}
%    \caption{Comparison score distribution plots of different datasets}
%    \label{fig:scores}
% \end{figure*}
\subsection{Histogram of Polynomial Coefficients}
We use three polynomial curves to mark the three main lines of the palmprint. 
Each polynomial curve contains five coefficients. 
Therefore, we plot the histograms for all 15 coefficients, as shown in Fig.\ref{fig:poly-hist}. 
Additionally, we conduct a statistical analysis of the x-coordinates for the endpoints of 3 palm lines. 
% 1 and Line 2, as well as the start point of Line 3.

% \subsection{Comparison Score Distributions}
% We plot genuine and imposter comparison score distributions for different datasets, including real datasets, PCE-Palm\cite{jin2024pce} generated, IDiff-Face\cite{boutros2023idiff} generated and Diff-Palm generated with varying $\gamma$, as shown in Fig.\ref{fig:scores}.
% Specifically, we use a recognition model pre-trained on the real dataset to extract features from each image in different datasets. Subsequently, we randomly select 100,000 genuine and imposter pairs, calculate their scores, and plot the results.

% \section{C. More Experimental Results}

\begin{figure}[!htb]
   \centering
   \small
   \center{\includegraphics[width=.95\linewidth] {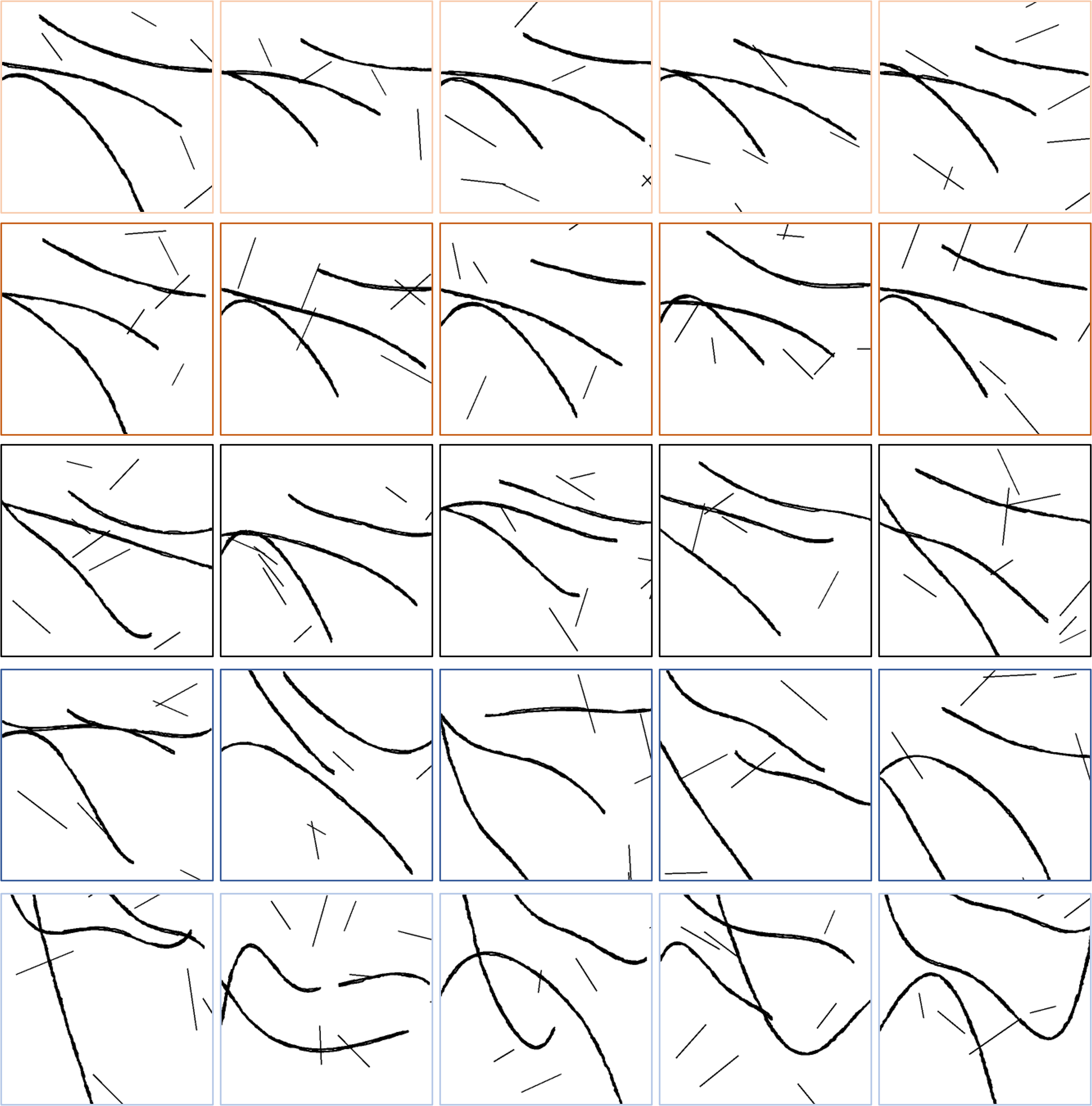}}
   \caption{Synthesized polynomial crease images with varying $\gamma$. From top to bottom, the $\gamma$ is set to $0.25$, $0.5$, $1.0$, $2.0$, and $4.0$, respectively}
   \label{fig:poly-sim}
\end{figure}

\begin{figure}[!htb]
   \centering
   \small
   \center{\includegraphics[width=.95\linewidth] {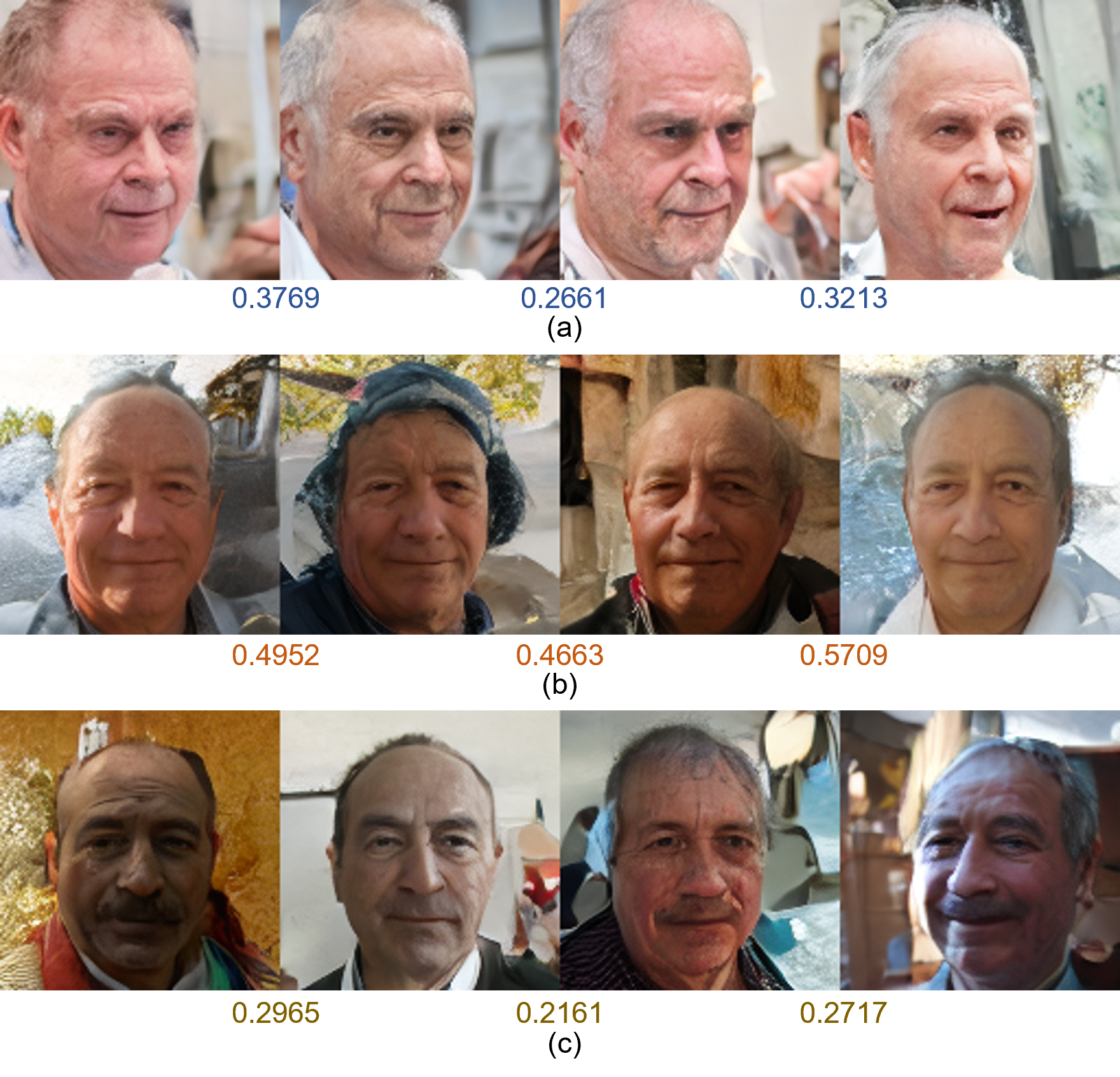}}
   \caption{Synthesized face images by IDiff-Face\cite{boutros2023idiff}, (a) with noise-sharing in first $500$ sampling steps, (b) with noise-sharing in last $500$ sampling steps, (c) without noise-sharing, 
   as well as cosine similarity calculated between adjacent face images}
   \label{fig:face}
\end{figure}

\begin{table*}[!htb]
    \centering
    \small
    
    \begin{tabular}{lccccccccc}
    \toprule
    \multirow{2}{*}{\textbf{Methods}} & \multicolumn{3}{c}{{\textbf{Configs}}} & \multicolumn{6}{c}{{\textbf{Performance (TAR@FAR=1e-6) $\uparrow$}}} \\
    \cmidrule(lr){2-4} \cmidrule(lr){5-10} 
    & \multirow{1}{*}{\#IDs} & \multirow{1}{*}{\#Images} & \multirow{1}{*}{FT w/ Real} &
    \multicolumn{1}{c}{CASIA} & \multicolumn{1}{c}{PolyU} & \multicolumn{1}{c}{TongJi} &
    \multicolumn{1}{c}{MPD} & \multicolumn{1}{c}{XJTU-UP} & \multicolumn{1}{c}{\textit{Avg.}} \\
    \midrule
    \foretab  CASIA & 310 & 2510 & \XSolidBrush     & 0.7243 & {0.7198}  & {0.5952} &{0.0964} & 0.1963 & {0.4664} \\
    \foretab PolyU & 194 & 3869 & \XSolidBrush     & 0.5235 & {0.7574}  & {0.4322} &{0.0746} & 0.1402 & {0.3856} \\
    \foretab TongJi & 300 & 6000 & \XSolidBrush    & 0.7344 & {0.7115}  & {0.8032} &{0.1476} & 0.0977 & {0.4989} \\
    \foretab MPD & 200 & 8000 & \XSolidBrush       & 0.8839 & {0.8254}  & {0.8599} &{0.3745} & 0.2939 & {0.6455} \\
    \foretab XJTU-UP & 100 & 3950 & \XSolidBrush   & 0.7741 & {0.7518}  & {0.7183} &{0.1805} & 0.4549 & {0.5759} \\
    \foretab IITD & 460 & 2601 &\XSolidBrush        & 0.3895 & {0.5205}  & {0.1024} &{0.0397} & 0.0681 & {0.2240} \\
    \foretab NTU-CP-v1 & 652 & 2390 &\XSolidBrush   & 0.6652 & {0.8330}  & {0.7970} &{0.1628} & 0.2588 & {0.5434} \\
    \midrule
    % Diff-Palm & 2k & 40k  & \Checkmark &{0.9787} &{0.9859} &{0.9744} &0.5979 & 0.8433 &{0.8760} \\
    Diff-Palm & 5k & 100k  & \Checkmark  & 0.9783 & 0.9850 & 0.9848 & 0.6754 & 0.8966 & 0.9040 \\
    Diff-Palm & 10k & 200k  & \Checkmark & 0.9857 & 0.9910 & 0.9905 & 0.7400 & 0.9442 & 0.9303 \\
    Diff-Palm & 20k & 400k  & \Checkmark & 0.9832 & 0.9943 & 0.9920 & 0.7922 & 0.9484 & 0.9420 \\
    Diff-Palm & 30k & 600k  & \Checkmark & 0.9870 & 0.9945 & 0.9941 & 0.8044 & 0.9570 & 0.9474 \\
    Diff-Palm & 40k & 800k  & \Checkmark & 0.9827 & 0.9949 & 0.9936 & 0.8179 & 0.9593 & 0.9497 \\ 
    Diff-Palm & 50k & 1M    & \Checkmark & 0.9843 & 0.9965 & 0.9933 & 0.8498 & 0.9700 & 0.9588 \\
    % \midrule
    % \rowcolor{gray!20} Diff-Palm(first $K$) & 2k & 40k & \XSolidBrush & 0.7649 & {0.7571}  & {0.5136} &{0.2502} & 0.0307 & {0.4633} \\
    % \rowcolor{gray!20} Diff-Palm(last $K$) & 2k & 40k & \XSolidBrush & 0.8782 & {0.9601}  & {0.9460} &{0.4643} & 0.6161 & {0.7729} \\
    \midrule
    %  \foretab Real data& 2.2k & 29.3k & \XSolidBrush & \textbf{0.9200} & 0.9196 & 0.9209 & 0.3877 & \textbf{0.6247} & 0.7546 \\
    % \foretab IDiff-Face\cite{boutros2023idiff} & 2k & 40k  & \XSolidBrush & 0.7977 & 0.7720 & 0.6115 & 0.1847 & 0.2667 & 0.5265 \\
    % % \rowcolor{gray!20} RPG-Palm\cite{shen2023rpg} & 2k &20 & 40k    & \XSolidBrush & - & - & - & - & - & - \\
    % \foretab PCE-Palm\cite{jin2024pce} & 2k  & 40k    & \XSolidBrush & 0.6796 & 0.7740 & 0.6149 & 0.2972 & 0.4150 & 0.5561 \\
    % \foretab Diff-Palm & 2k  & 40k        & \XSolidBrush & 0.8782 & \textbf{0.9601} & \textbf{0.9460} & \textbf{0.4643} & 0.6161 & \textbf{0.7729} \\
    % \midrule
    \foretab Diff-Palm(10k) & 2k & 40k & \XSolidBrush & 0.8598 & {0.9472}  & {0.9113} &{0.4124} & 0.6133 & {0.7488} \\
    \foretab Diff-Palm(48k) & 2k & 40k & \XSolidBrush & 0.8782 & {0.9601}  & {0.9460} &{0.4643} & 0.6161 & {0.7729} \\
    \midrule
     Real data(R50) & 2.2k & 29.3k                      & \XSolidBrush & 0.9429 & 0.8927 & 0.9449 & 0.3974 & 0.6111 & 0.7577 \\
     IDiff-Face(R50) \cite{boutros2023idiff} & 2k & 40k  & \XSolidBrush & 0.7944 & 0.7404 & 0.6208 & 0.1793 & 0.2734 & 0.5217 \\
     PCE-Palm(R50) \cite{jin2024pce} & 2k & 40k          & \XSolidBrush & 0.5749 & 0.7188 & 0.5835 & 0.2738 & 0.4347 & 0.5171 \\
     Diff-Palm(R50) & 2k & 40k                          & \XSolidBrush & 0.8708 & 0.9558 & 0.9577 & 0.4472 & 0.6266 & 0.7716 \\
    \midrule
    
    \foretab Real data(MBF) & 2.2k & 29.3k                      & \XSolidBrush & 0.9323 & 0.9071 & 0.9247 & 0.3616 & 0.6067 & 0.7466 \\
    \foretab IDiff-Face(MBF) \cite{boutros2023idiff} & 2k & 40k  & \XSolidBrush & 0.7732 & 0.7445 & 0.6507 & 0.1667 & 0.2259 & 0.5120 \\
    \foretab PCE-Palm(MBF) \cite{jin2024pce} & 2k & 40k          & \XSolidBrush & 0.6112 & 0.7379 & 0.5266 & 0.2695 & 0.4013 & 0.5093 \\
    \foretab Diff-Palm(MBF) & 2k & 40k                          & \XSolidBrush & 0.8546 & 0.9489 & 0.9409 & 0.4297 & 0.6177 & 0.7584 \\
    \midrule
    Real data(ViT) & 2.2k & 29.3k                      & \XSolidBrush     & \textbf{0.8279}    & 0.6909    & 0.7390    & 0.2132    & 0.2826    & 0.5505 \\
    
    IDiff-Face(ViT)  \cite{boutros2023idiff} & 2k & 40k  & \XSolidBrush & 0.6676    & 0.5968    & 0.5304    & 0.1234    & 0.1783    & 0.4193  \\
    
     Vec2Face(ViT)  \cite{wu2024vec2face} & 2k & 40k  & \XSolidBrush& 0.6967    & 0.5075    & 0.4406    & 0.0962    &  0.1108    & 0.3704 \\
     
     PCE-Palm(ViT) \cite{jin2024pce} &  2k & 40k  & \XSolidBrush   & 0.6012    & 0.4918    & 0.4604    & 0.1303    & 0.1515    & 0.3670  \\
     Diff-Palm(ViT)       & 0.6814    & \textbf{0.7920}    & \textbf{0.7792}    & \textbf{0.2798}    & \textbf{0.3754}    & \textbf{0.5816} \\
    \bottomrule
    
    \end{tabular}
    \caption{Comparsion performance of recognition models trained on various datasets. Results are reported in TAR@FAR=$1e-6$. `R50', `MBF' and `ViT' represent ResNet-50\cite{deng2019arcface}, MobileFaceNet\cite{chen2018mobilefacenets} and ViT-t \cite{Dosovitskiy2020AnII}, respectively}
    \label{tab:sup}
    
\end{table*}

\subsection{Performance on Individual Public Datasets}
We conduct recognition experiments on individual public datasets. 
The experimental results are presented in the first section of Tab.\ref{tab:sup}. 
The performance achieved on individual public datasets is significantly lower compared to that on mixed public datasets.

% \subsection{Number of Synthesized Samples per Identity}
% Using our proposed Diff-Palm, we generate datasets maintaining 2k IDs, each with varying numbers of samples. 
% The results of training the recognition model on these datasets are shown in the second section of Tab.\ref{tab:sup}.
% We observe that as the number of samples per ID increases, the performance of the recognition model continuously improves. 
% The optimal recognition results are achieved when generating 200 samples per ID.

\subsection{Further Fine-Tuning Experiments}
We adopt Diff-Palm to generate datasets with a large number of IDs. 
These datasets are used to pre-train the recognition model, which is subsequently fine-tuned using real datasets. 
As shown in the second section of Tab.\ref{tab:sup}, the performance of our method consistently improves after fine-tuning.

\subsection{Polynomial Creases Similarity Control}
We generate a polynomial creases dataset and control the overall similarity using different $\gamma$. 
As illustrated in Fig.\ref{fig:poly-sim}, we can observe that when $\gamma$ is less than $1.0$, the similarity of the generated polynomial creases increases. 
Conversely, when $\gamma$ is greater than $1.0$, the overall similarity decreases, and the generated creases become more random.

\subsection{Additional Ablation Experiments}
% \textbf{$K$-Step Noise-Sharing Sampling}
% In our paper, we have discussed two approaches for applying $K$-step noise-sharing. 
% The noise-sharing can be applied in either the first $K$ steps or the last $K$ steps during the sampling process, leading to different outcomes. 
% By default, we use the latter approach. 
% Comparative results with the former approach are presented in the fifth section of Tab.\ref{tab:sup}. 
% It is evident that using the first $K$ results in poor recognition performance.

\textbf{Smaller Anonymous Datasets}
We have collected an anonymous dataset containing 48,000 images, which we used to train the generative model. 
We also experiment with training Diff-Palm using a smaller dataset of 10,000 images. 
The experimental results are presented in the third section of Tab.\ref{tab:sup}. 
We observe that the performance of Diff-Palm trained with 10,000 images is inferior to that trained with 48,000 images. 
However, it still achieves results comparable to those obtained with real datasets.

\textbf{Different Recognition Backbone}
We conduct comparative experiments using different recognition backbones (modified Resnet-50 \cite{deng2019arcface}, MobileFaceNet \cite{chen2018mobilefacenets} and ViT-t \cite{Dosovitskiy2020AnII} ). 
The experimental results are shown in the last three sections of Tab.\ref{tab:sup}. 
We arrive at the same conclusions as those in the main paper. 

% We have collected an anonymous dataset containing 48,000 images, which we used to train the generative model. 
% We also experiment with training Diff-Palm using a smaller dataset of 10,000 images. 
% The experimental results are presented in the last section of Tab.\ref{tab:sup}. 
% We observe that the performance of Diff-Palm trained with 10,000 images is inferior to that trained with 48,000 images. 
% However, it still achieves results comparable to those obtained with real datasets.

\section{More Discussion}

\subsection{\textit{K}-Step Noise-Sharing Sampling}
To verify that $K$-step noise-sharing sampling can be applied to other diffusion-based methods, we use the officially released pre-trained IDiff-Face model \cite{boutros2023idiff} and apply our $K$-step noise-sharing sampling to obtain facial images, as shown in Fig.\ref{fig:face}. 
We set $K=500$ with a total step of  $T=1000$. 
Each row of images is generated from the same ID condition. 
Moreover, we employ a pre-trained facial recognition model to extract features from each image and 
calculate the cosine similarity between adjacent face images. 
It is evident that applying $K$-step noise-sharing sampling significantly enhances the identity consistency of the generated results. 
Additionally, applying noise-sharing in the last $K$ steps further improves the identity consistency of the generated outcomes.

% When applying noise-sharing in the first $K$ steps, the generated results exhibit the same pose but different facial expressions, as shown in Fig.\ref{fig:face}(a). 
% Conversely, when applying noise-sharing in the last $K$ steps, the generated results exhibit the same facial expression but different poses, as shown in Fig.\ref{fig:face}(b). 
% However, Without using noise sharing, controlling the consistency of these attributes in the generated results is challenging.

\subsection{Validation Set}
In PCE-Palm, they first split several public datasets into training and testing sets in a 1:1 ratio and then mixed all the testing sets from the public datasets for evaluation with the trained recognition model. 
However, due to different collection devices, environments, etc., the various public datasets have significant style differences. 
When the recognition model is tested on the mixed testing set, it is easy to distinguish an identity from one dataset from identities in other datasets. 
In contrast, we adopt a more general validation approach. 
After splitting the public data into training and testing sets in a 1:1 ratio, we validate each dataset separately. 
Subsequently, we average the validation results from each dataset.

% \section{Rationale}
% \label{sec:rationale}
% % 
% Having the supplementary compiled together with the main paper means that:
% % 
% \begin{itemize}
% \item The supplementary can back-reference sections of the main paper, for example, we can refer to \cref{sec:intro};
% \item The main paper can forward reference sub-sections within the supplementary explicitly (e.g. referring to a particular experiment); 
% \item When submitted to arXiv, the supplementary will already included at the end of the paper.
% \end{itemize}
% % 
% To split the supplementary pages from the main paper, you can use \href{https://support.apple.com/en-ca/guide/preview/prvw11793/mac#:~:text=Delete%20a%20page%20from%20a,or%20choose%20Edit%20%3E%20Delete).}{Preview (on macOS)}, \href{https://www.adobe.com/acrobat/how-to/delete-pages-from-pdf.html#:~:text=Choose%20%E2%80%9CTools%E2%80%9D%20%3E%20%E2%80%9COrganize,or%20pages%20from%20the%20file.}{Adobe Acrobat} (on all OSs), as well as \href{https://superuser.com/questions/517986/is-it-possible-to-delete-some-pages-of-a-pdf-document}{command line tools}.

\end{document}